\documentclass[10pt, letter, onecolumn]{arxiv}
\usepackage{hyperref}
\usepackage{caption}
\hypersetup{
  pagebackref,
  breaklinks,
  colorlinks
}
\usepackage{lineno}
\usepackage{booktabs}   
\usepackage{threeparttable}
\usepackage{makecell}
\usepackage{bbding}
\usepackage{mdframed}
\usepackage{tabulary}
\usepackage{colortbl}
\usepackage{tcolorbox}
\usepackage{utfsym}
\usepackage{tikz}
\usepackage{soul}
\usepackage{booktabs}
\usepackage{graphicx}   
\usepackage[hang,flushmargin]{footmisc}
\usepackage[dvipsnames,table]{xcolor}

\usepackage{graphicx}%
\usepackage{multirow}%
\usepackage{amsmath,amssymb,amsfonts}%
\usepackage{amsthm}%
\usepackage{mathrsfs}%
\usepackage[title]{appendix}%
\usepackage{xcolor}%
\usepackage{textcomp}%
\usepackage{manyfoot}%
\usepackage{booktabs}%
\usepackage{algorithm}%
\usepackage{rotating}%
\usepackage{algorithmicx}%
\usepackage{algpseudocode}%
\usepackage{listings}%
\usepackage{enumitem}
\usepackage{hyperref}
\usepackage[T1]{fontenc}
\usepackage{ulem}

\graphicspath{{figures/}}

\definecolor{lightskyblue}{rgb}{0.53, 0.81, 0.98}
\definecolor{lightgray}{RGB}{242,242,242}
\definecolor{lightbrown}{RGB}{242,225,216}
\definecolor{lightblue}{RGB}{246,248,252}

\colorlet{"0-7"}{GreenYellow}
\colorlet{"7-14"}{SpringGreen}
\colorlet{"14-"}{LimeGreen}

\colorlet{"0-2"}{GreenYellow}
\colorlet{"2-4"}{SpringGreen}
\colorlet{"4-"}{LimeGreen}
\definecolor{"bad"}{RGB}{246,161,147}

\definecolor{lightlightgray}{rgb}{200,200,200}

\newcommand{\genmodel}[1]{{MedReCo-VLM}}
\newcommand{\retmodel}[1]{{MedReCo}}
\newcommand{\dataset}[1]{{MedReCo-DB}}
\title{\Large{A Vision–language Framework for Comparative Reasoning \\ in Radiology}}

\vspace{1cm}

\author[1,3,*]{Tengfei Zhang}
\author[2,3,*]{Ziheng Zhao}
\author[4,*]{Xiaoman Zhang}
\author[5,6]{Lisong Dai}
\author[2,3]{Pengcheng Qiu}
\author[2,3]{Ya Zhang}
\author[2,3,$\dag$]{Yanfeng Wang}
\author[2,3,$\dag$]{Weidi Xie}

\affil[1]{\small University of Science and Technology of China \authorcr} 
\affil[2]{\small School of Artificial Intelligence, Shanghai Jiao Tong University \authorcr} 
\affil[3]{\small Shanghai Artificial Intelligence Laboratory \authorcr} 
\affil[4]{\small Department of Biomedical Informatics, Harvard Medical School \authorcr}
\affil[5]{\small Department of Radiology, Renmin Hospital of Wuhan University \authorcr} 
\affil[6]{\small Shanghai Sixth People’s Hospital Affiliated to Shanghai Jiao Tong University \authorcr} 

\renewcommand{\correspondingauthor}[1]{$\dag$~Corresponding Authors.}

\begin{document}
\begin{abstract}

\textbf{Abstract.}
Medical imaging artificial intelligence has achieved strong performance in isolated image interpretation, but remains poorly aligned with radiological practice, where diagnosis and follow-up rely on comparison across prior studies and analogous reference cases. Here we formulate radiological comparison as an entity-aware cross-image reasoning problem and introduce a framework that supports both reference-case retrieval and temporal comparative interpretation. We construct \textbf{MedReCo-DB}, a large-scale comparative imaging resource derived from routine image–report pairs, comprising more than 690,000 images from over 160,000 patients across eight institutions, four countries and seven imaging modalities. Reports are decomposed into anatomical structures, abnormal findings and pathological conditions to provide supervision for entity-conditioned retrieval and comparative visual question answering.
Using this resource, we develop \textbf{MedReCo}, an entity-aware visual encoder for controllable retrieval of clinically analogous cases, and \textbf{MedReCo-VLM}, a vision--language extension for generative interpretation of interval change.
Across internal, external and cross-center evaluations, \textbf{MedReCo} achieved the highest Recall@1 in all 12 internal retrieval settings and improved external retrieval by a mean of 6.0 percentage points. 
In clinically confusable differential groups, it achieved positive Recall@1 gains over the strongest baselines.
\textbf{MedReCo-VLM} achieved the best performance across all comparative generation evaluations and improved longitudinal follow-up accuracy by 14.5-46.5 percentage points on chest radiographs and 13.0–27.9 percentage points on CT. 
These findings suggest that entity-aware comparative reasoning can be learned from routine clinical data at scale and may provide a more clinically aligned foundation for medical imaging AI.

\end{abstract}
\maketitle



\section{Introduction}

\begin{figure}[p]
  \centering
  \includegraphics[width=\linewidth]{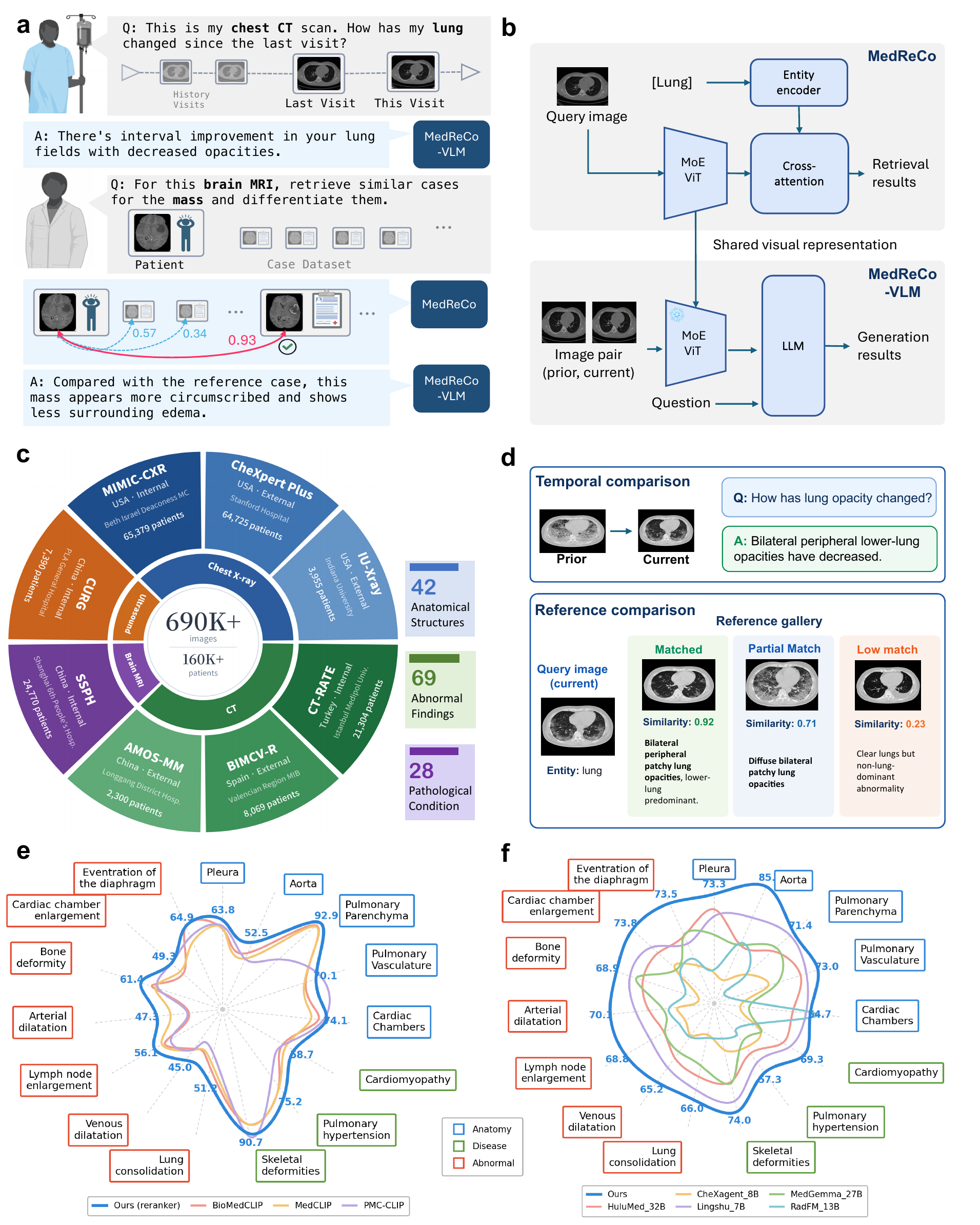}
  \vspace{1pt}
  \caption{
    \textbf{Overview of this study.}
    \textbf{a,} Radiological comparison is formulated around temporal follow-up and reference-case reasoning.
    \textbf{b,} MedReCo uses an entity-conditioned visual representation to support both retrieval and vision--language generation.
    \textbf{c,} MedReCo-DB integrates multi-modal imaging data with entity-level clinical annotations.
    \textbf{d,} The benchmark includes both prior--current temporal comparisons and cross-case reference comparisons.
    \textbf{e,} Representative retrieval results evaluate entity-conditioned reference matching.
    \textbf{f,} Representative VQA results evaluate entity-specific comparative reasoning.
    }
  \vspace{-.4cm}
  \label{fig:overview}
\end{figure}

Artificial intelligence has achieved strong performance across a wide range of medical image-analysis tasks, including radiology report generation~\cite{jing2018automatic,li2019knowledge}, image segmentation~\cite{isensee2021nnu,ma2024segment,zhao2025large}, disease classification~\cite{irvin2019chexpert,rajpurkar2017chexnet} and visual question answering~\cite{RADFM,luo2023biomedgpt,jiang2025hulu}. Yet most medical imaging AI systems still operate under a single-image paradigm: they interpret one examination at a time, often without explicit access to prior studies or analogous reference cases. This differs fundamentally from radiological practice, where images are rarely interpreted in isolation.  
Instead, radiologists ask comparative questions throughout the diagnostic process: What does this finding resemble? Has it appeared before? Has it grown, resolved or changed in distribution? The ability to compare images across patients and across time is therefore not a peripheral function, but a central component of clinical image interpretation~\cite{eisenhauer2009recist,roelofs2007prior,wang2026ai} (Fig.~\ref{fig:overview}a).

We frame comparative reasoning in radiology around two complementary settings that recur across clinical workflows. 
The first is \textbf{reference comparison}, in which a current finding is interpreted by comparison with similar historical, canonical or previously diagnosed cases~\cite{akgul2011content,choe2022content,muller2004cbir}. This process supports differential diagnosis, helps distinguish visually confusable entities and reflects the way radiologists draw on accumulated experience, institutional archives and the literature. The second is \textbf{temporal comparison}, in which a current examination is compared with prior studies from the same patient to assess disease progression, treatment response, stability, or resolution~\cite{eisenhauer2009recist,roelofs2007prior}. These two forms of comparison serve different clinical purposes, but they share a common computational requirement: given a target clinical entity, the model must identify the corresponding or analogous entity in another image and reason about its similarity, difference or interval change. These two comparison settings are instantiated by the representative temporal and reference-comparison examples shown in Fig.~\ref{fig:overview}d.

Clinical comparison is inherently entity-specific: the same pair of studies may be similar with respect to one entity but different with respect to another. For example, two chest CT scans may have globally similar thoracic anatomy while differing in pulmonary arterial enlargement, hilar lymphadenopathy or ground-glass opacity. Conversely, two longitudinal examinations may show the same disease category but differ in lesion size, distribution or internal composition. A clinically useful comparative model must therefore move beyond global image-level matching. It must represent fine-grained anatomical structures, abnormal findings and pathological conditions; align these entities across images; and express clinically meaningful similarities, differences and temporal changes in language.

Current medical AI systems only partially address this need. Image retrieval models can identify visually similar cases, but most rely on holistic image embeddings and are not controllable by a specific clinical entity~\cite{lin2023pmc,wang2022medclip,zhang2023biomedclip,hamamci2026generalist}. Medical vision–language models can generate reports or answer questions, but they are commonly trained on single-image image–text alignment and often lack explicit supervision for cross-image, entity-level comparison~\cite{huang2021gloria,bannur2024maira2,RADFM,xu2025lingshu,jiang2025hulu,sellergren2025medgemma}. Pairwise or longitudinal report-generation models have begun to address temporal interpretation, but they are usually developed for narrow modalities or tasks and are not integrated with reference-case retrieval~\cite{dalla2023controllable,zhu2023utilizing,wang2024hergen}. Progress has also been limited by the absence of large-scale datasets that provide structured supervision for comparative reasoning across modalities, institutions, and clinical entities~\cite{hu2023expert,prakash2026chextemporal,gai20263d,li2026radthinking}.

Here, we formulate radiological comparison as an entity-aware cross-image reasoning problem and introduce a framework designed to support two clinically central workflows: controllable reference-case retrieval and generative temporal comparative interpretation (Fig.~\ref{fig:overview}b). 
The framework is built around a shared visual representation that is conditioned on clinical entities, including anatomical structures, abnormal findings and pathological conditions. This design allows the model to compare images with respect to a specified entity rather than only according to global visual appearance. The retrieval component, \textbf{MedReCo}, ranks candidate studies according to entity-specific similarity, enabling reference comparison for differential diagnosis. The generative component, \textbf{MedReCo-VLM}, adapts the same entity-aware visual representation to a large language model, enabling natural-language descriptions of similarities, differences and interval changes between paired studies.

To develop and evaluate this framework, we constructed MedReCo-DB, a large-scale comparative medical imaging resource derived from routine image–report pairs. MedReCo-DB comprises more than 690,000 images from over 160,000 patients across eight institutions, four countries and seven imaging modalities, including chest radiography, chest CT, abdominal CT, brain MRI, and breast, thyroid and liver ultrasound (Fig.~\ref{fig:overview}c). 
The dataset was built on the premise that routine radiology reports contain rich entity-level information that can be converted into scalable supervision for comparative reasoning. We decomposed reports into structured descriptions of anatomical structures, abnormal findings and pathological conditions. 
These entity-specific descriptions were then used to construct two complementary forms of supervision: ranking triplets for entity-conditioned retrieval and comparative visual question-answering examples for generative interpretation (Fig.~\ref{fig:overview}d).
The resulting ontology covers 42 anatomical structures, 69 abnormal findings and 28 pathological conditions across the seven imaging modalities.

MedReCo uses this supervision to learn fine-grained, entity-aware visual representations through text-guided contrastive ranking. A modality-aware vision encoder accommodates the heterogeneity of radiography, CT, MRI and ultrasound, while entity-conditioned attention helps isolate the visual evidence relevant to the queried clinical concept. MedReCo-VLM then connects the pretrained visual encoder to a large language model through instruction tuning, allowing the system to generate comparative interpretations for image pairs. By sharing the same visual foundation across retrieval and generation, the framework links reference comparison and temporal comparison as two manifestations of a common entity-level reasoning problem.

Representative entity-level retrieval and comparative VQA results are highlighted in Fig.~\ref{fig:overview}e,f, with comprehensive evaluations reported in subsequent result sections. We evaluated MedReCo and MedReCo-VLM across progressively more demanding settings. For reference comparison, we assessed internal retrieval, external validation on held-out institutions, cross-center retrieval and clinically confusable differential-diagnosis groups. MedReCo achieved the highest Recall@1 in all 12 internal retrieval settings, improved external Recall@1 by a mean of 6.0 percentage points and consistently outperformed the strongest baselines in clinically confusable differentials. For temporal comparison, we evaluated broad pairwise comparative interpretation, independent public paired-image VQA benchmarks and same-patient longitudinal follow-up studies. MedReCo-VLM achieved the best performance across all 24 comparative interpretation evaluations, reached 87.1\% accuracy on public paired-image VQA benchmarks and improved longitudinal follow-up accuracy by 14.5-46.5 percentage points on chest radiographs and by 13.0–27.9 percentage points on CT. Together, these results suggest that entity-aware comparative reasoning can be learned from routine clinical data at scale and may provide a more clinically aligned foundation for medical imaging AI.

\section{Results}

We evaluated \retmodel~ and \genmodel~ on entity-conditioned tasks designed to instantiate comparative radiological reasoning. 
For \textbf{reference comparison}, the {controllable image retrieval} task takes a query image and a target clinical entity, and ranks candidate images according to entity-specific similarity to retrieve clinically analogous reference cases. 
For \textbf{temporal comparison}, the {generative comparative interpretation} task takes an image pair and a target clinical entity, and generates a clinically meaningful description of similarities, differences, or interval changes. Both tasks were evaluated at three levels of clinical granularity: anatomical structures, abnormal findings, and pathological conditions.

\textbf{Evaluation datasets.} 
\dataset{} aggregates report-paired images from seven public datasets and one in-house collection, 
spanning four countries (the USA, Turkey, Spain, and China) and seven imaging modalities: brain MRI, chest X-ray, chest CT, abdominal CT, and breast, thyroid and liver ultrasound.
For internal validation, we used four public datasets (MIMIC-CXR~\cite{johnson2019mimic}, CT-RATE~\cite{hamamci2026generalist}, AMOS-MM~\cite{ji2022amos}, and CURG~\cite{li2024ultrasound}), following their official train-test splits, along with an in-house brain MRI dataset~\cite{LEI2025102516} split 9:1 for training and testing. For external validation, we evaluated the model on held-out datasets, including IU-Xray~\cite{IUXray}, CheXpert Plus~\cite{chambon2024chexpert}, and BIMCV-R~\cite{chen2024bimcv}. 
To further assess comparative interpretation, we additionally evaluated on Medical-Diff-VQA~\cite{PhysioNet-medical-diff-vqa-1.0.1} and MMXU~\cite{mu2025mmxu}, two independent public paired-image VQA benchmarks focused on chest radiographs. More dataset information are provided in Section~\ref{sec:data_sources}, \ref{sec:entity_definition} and \ref{sec:data_construction}, while the benchmark composition is detailed in Table~\ref{tab:benchmark_retrieval_composition}~\ref{tab:benchmark_vqa_composition}~\ref{tab:supp-modality}.


\subsection{Reference comparison with controllable image retrieval}

We first evaluated the reference-comparison capability of \retmodel~, using \textit{controllable image retrieval} across diverse medical image archives. 
For each query comprising an image and a specified clinical entity, \retmodel~ ranked candidate images by entity-specific similarity. 
This setting requires the retrieval of cases that match the queried entity, rather than merely sharing modality, anatomical region, or overall visual appearance. 
Performance was measured using $\mathrm{Recall}@k$ (\(k \in \{1,3,5\}\)) under an entity-level relevance criterion (Section~\ref{sec:data_construction}). 
We compared \retmodel~ with five baseline models: CT-CLIP~\cite{hamamci2026generalist}, MedCLIP~\cite{wang2022medclip}, PMC-CLIP~\cite{lin2023pmc}, BiomedCLIP~\cite{zhang2023biomedclip}, and HLIP~\cite{zhao2025towards}. 
Each baseline was evaluated only on the imaging modalities for which it was designed.
As these baselines do not support entity-conditioned retrieval, we evaluate them using similarity scores and rankings derived from holistic image whole-image features.
Results are reported across four settings of increasing stringency: internal validation, external validation, cross-center retrieval, and clinically confusable differentials.

\begin{figure}[!t]
  \centering
  \includegraphics[width=\linewidth]{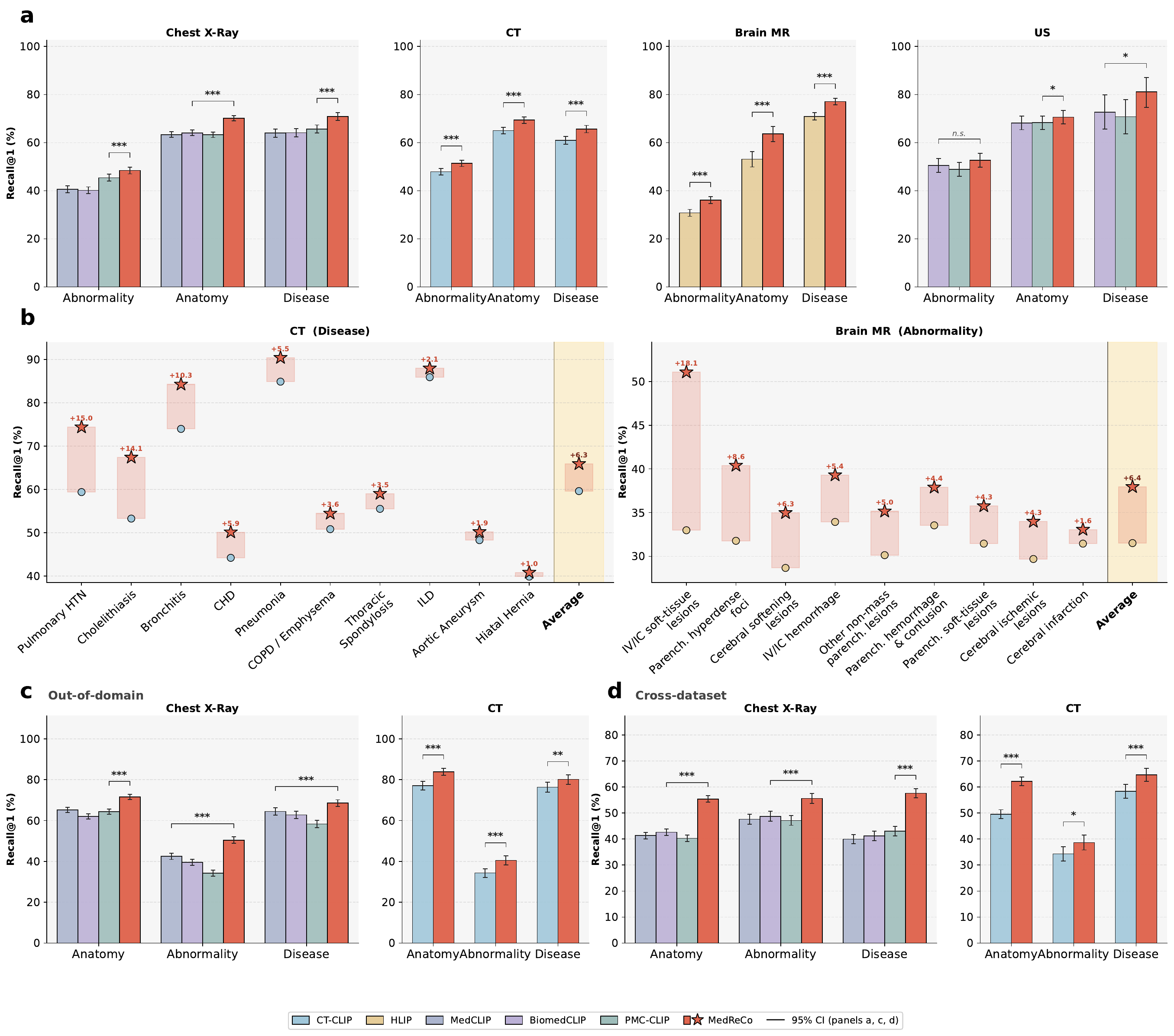}
  \vspace{1pt}
    \caption{
    \textbf{Entity-conditioned reference-case retrieval performance.}
    \textbf{a,} Internal-validation results across four imaging modalities and three entity levels: anatomical structures, abnormal findings and pathological conditions. Recall@1 is reported. Baselines are shown only for their intended modalities. Statistical significance is denoted as *\textit{P}<0.05, **\textit{P}<0.01 and ***\textit{P}<0.001; n.s., not significant.
    \textbf{b,} Abnormal-finding subcategory analysis in CT and brain MR. Star markers denote \retmodel~ and circular markers denote baseline methods. All comparisons in this panel meet the significance threshold \textit{P}<0.001. Abbreviations: ILD, interstitial lung disease; CHD, coronary heart disease; HTN, hypertension; Parench., parenchymal; IV/IC, intraventricular/intracisternal.
    \textbf{c,} External validation results on held-out chest-X-ray and CT cohorts.
    \textbf{d,} Cross-center retrieval from external reference archive.
    }
  \vspace{-.4cm}
  \label{fig:retrieval_results}
\end{figure}

\textbf{Internal validation.} 
We first evaluated retrieval in an internal setting, where both queries and candidate studies were drawn from the same institution and imaging modality, and all cases were represented in the training distribution.
\retmodel~ achieved the highest $\mathrm{Recall}@1$ in all 12 internal retrieval settings across four imaging modalities and three entity levels (Figure~\ref{fig:retrieval_results}a; Supplementary Table~\ref{tab:supp_retrieval_internal_compact}). 

Across all imaging modalities, \retmodel~ improved Recall@1 over the strongest applicable baseline by 5.8, 3.5, and 6.1 percentage points for anatomical structures, abnormal findings, and pathological conditions, respectively. 
We further stratified performance by individual abnormality categories in CT and brain MRI (Figure~\ref{fig:retrieval_results}b; Supplementary Table~\ref{tab:supp_retrieval_finegrained_compact}). 
Across 10 CT disease entities, \retmodel~ outperformed CT-CLIP by a mean of 6.3 percentage points in Recall@1. 
In brain MRI, \retmodel~ improved mean Recall@1 from 31.5\% with HLIP to 38.0\%. 
These gains were observed across most abnormality categories, indicating that the improvement was broadly distributed rather than driven by a small subset of findings. 

\begin{figure}[!t]
  \centering
  \includegraphics[width=\linewidth]{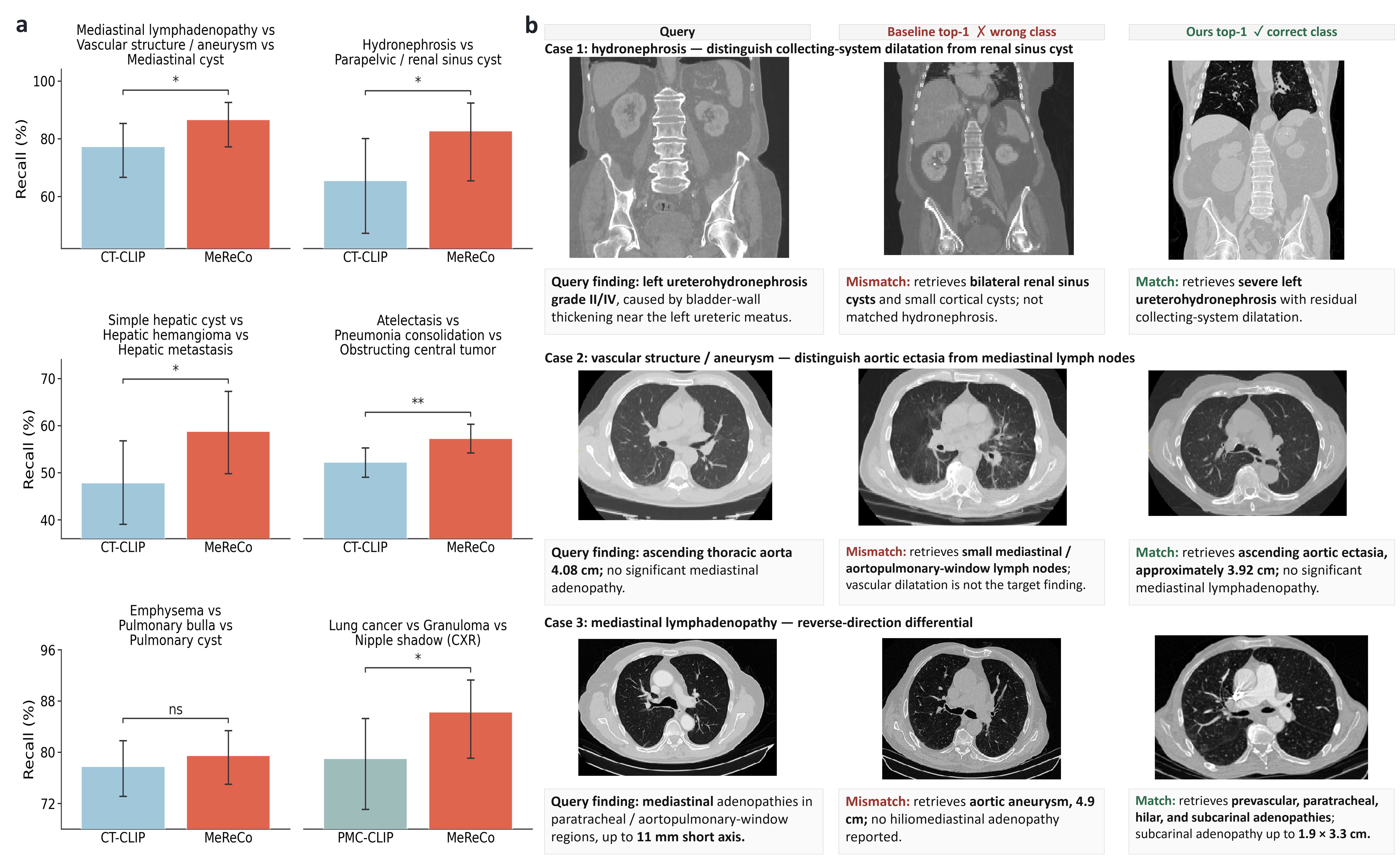}
  \vspace{2pt}
  \caption{
    \textbf{Controllable image retrieval on clinically confusable differentials.}
    \textbf{a,} Recall@1 on six visually confusable CT and chest-X-ray differential groups, where candidates share the same modality and anatomical region. 
    Random indicates chance performance; 
    CT-CLIP and PMC-CLIP are the strongest applicable CT and chest-X-ray baselines. 
    \textbf{b,} Representative retrieval examples for pulmonary-artery enlargement versus hilar lymphadenopathy on CT studies. 
  }
  \vspace{-.4cm}
  \label{fig:fine_retrieval_results}
\end{figure}

\textbf{External validation within held-out centers.}
To assess robustness to institutional shift, we next evaluated \retmodel~ on held-out datasets from previously unseen clinical centers (Figure~\ref{fig:retrieval_results}c; Supplementary Table~\ref{tab:supp_retrieval_external_crossdataset_compact}). In this setting, both query and candidate studies were drawn from the same held-out institution. 
\retmodel~ achieved 71.5\% Recall@1 for anatomical-structure retrieval on external chest radiographs, exceeding PMC-CLIP by 7.2 percentage points, and 83.9\% on external CT, exceeding CT-CLIP by 6.8 percentage points. 
These gains were observed across all six external chest-radiograph and CT settings, indicating robustness to differences in acquisition protocols, reporting conventions and patient populations.

\textbf{Cross-center retrieval from external institutions.}
We then evaluated a stricter cross-center setting in which queries and candidate galleries were drawn from non-overlapping institutions (Figure~\ref{fig:retrieval_results}d; Supplementary Table~\ref{tab:supp_retrieval_external_crossdataset_compact}). 
Specifically, MIMIC-CXR test images were queried against galleries from CheXpert Plus and IU-Xray, and CT-RATE test volumes were queried against BIMCV-R. 
This setting tests whether clinically relevant reference cases can be retrieved from entirely new institutional archives. 
\retmodel~ achieved the highest Recall@1 across all six modality-granularity combinations, exceeding PMC-CLIP by 11.4 percentage points on chest radiographs and CT-CLIP by 7.8 percentage points on CT. 
The largest gain was observed for disease-based chest-radiograph retrieval (+14.6 percentage points). 
These findings suggest that \retmodel~ can retrieve clinically relevant reference cases across institutions with reduced reliance on center-specific visual or reporting patterns.

\textbf{Retrieval for clinically confusable differentials.} 
To test performance in more clinically demanding settings, we curated six groups of confusable differential diagnoses from the external cohorts, each containing two or three findings arising in the same modality and anatomical context but requiring different diagnostic interpretations (Section~\ref{sec:data_construction}; Supplementary Table~\ref{tab:supp_differentials_compact}). 
In these scenarios, visually similar findings may correspond to distinct disease processes or management pathways, making accurate reference retrieval particularly informative. 
\retmodel~ improved Recall@1 over the strongest baseline in all six groups, with a macro-average gain of 8.6 percentage points (Figure~\ref{fig:fine_retrieval_results}a; Supplementary Table~\ref{tab:supp_differentials_compact}). 
The largest gains were observed for \textit{hydronephrosis} versus \textit{parapelvic/renal sinus cyst} (+17.2 percentage points) and \textit{simple hepatic cyst} versus \textit{hepatic hemangioma} versus \textit{hepatic metastasis} (+10.9 percentage points).
The remaining groups, including mediastinal lymphadenopathy versus vascular structure/aneurysm versus mediastinal cyst, atelectasis versus pneumonia consolidation versus obstructing central tumor, emphysema versus pulmonary bulla versus pulmonary cyst, and lung cancer versus granuloma versus nipple shadow on chest X-ray, also showed consistent gains, ranging from +1.7 to +9.3 percentage points.
These results indicate that \retmodel~ captures subtle but clinically consequential distinctions relevant to differential diagnosis.

\textbf{Representative retrieval examples.} Representative examples across clinically confusable CT differentials illustrate the difference between global visual similarity and entity-matched retrieval (Figure~\ref{fig:fine_retrieval_results}b). For a query targeting left ureterohydronephrosis, the baseline retrieved a visually plausible case of bilateral renal sinus cysts and small cortical cysts, whereas MedReCo retrieved a case with matched severe left ureterohydronephrosis and residual collecting-system dilatation. For a query targeting ascending aortic ectasia, the baseline retrieved a case with small mediastinal and aortopulmonary-window lymph nodes rather than vascular dilatation, whereas MedReCo retrieved a case with matched ascending aortic ectasia of similar calibre. For a query targeting mediastinal lymphadenopathy, the baseline retrieved a case of aortic aneurysm without reported hilar or mediastinal lymphadenopathy, whereas MedReCo retrieved a case characterized by prevascular, paratracheal, hilar and subcarinal lymphadenopathy. Across these examples, the baseline retrieved CT studies with broadly similar appearance or anatomical context, but failed to identify clinically relevant cases with respect to the specific, localized clinical entity of interest.

\begin{figure}[!t]
  \centering
  \includegraphics[width=\linewidth]{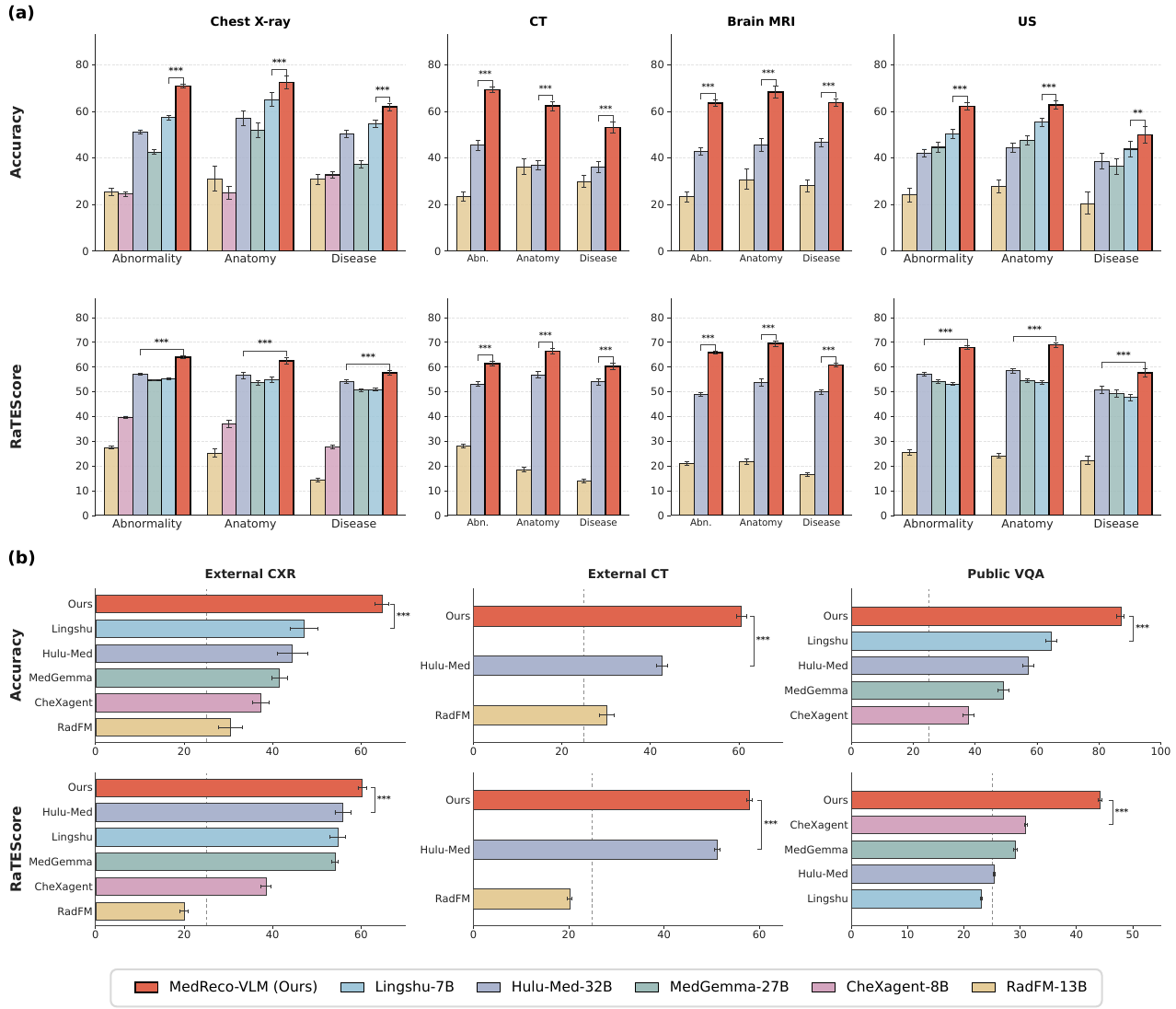}
  \vspace{2pt}
  \caption{
    \textbf{Generative comparative interpretation across cross-patient image pairs.}
    \textbf{a,} Internal validation result of generative comparative interpretation across four imaging modalities and three entity levels. Accuracy and RaTEScore are reported for close-ended and open-ended question respectively. Statistical significance is denoted as *\textit{P}<0.05, **\textit{P}<0.01 and ***\textit{P}<0.001; n.s., not significant. 
    \textbf{b,} Results of external validation and independent public benchmarks. The dashed line denotes the 25\% random reference.
    }
  \vspace{-.4cm}
  \label{fig:comparison_results}
\end{figure}

\begin{figure}[!t]
  \centering
  \includegraphics[width=\linewidth]{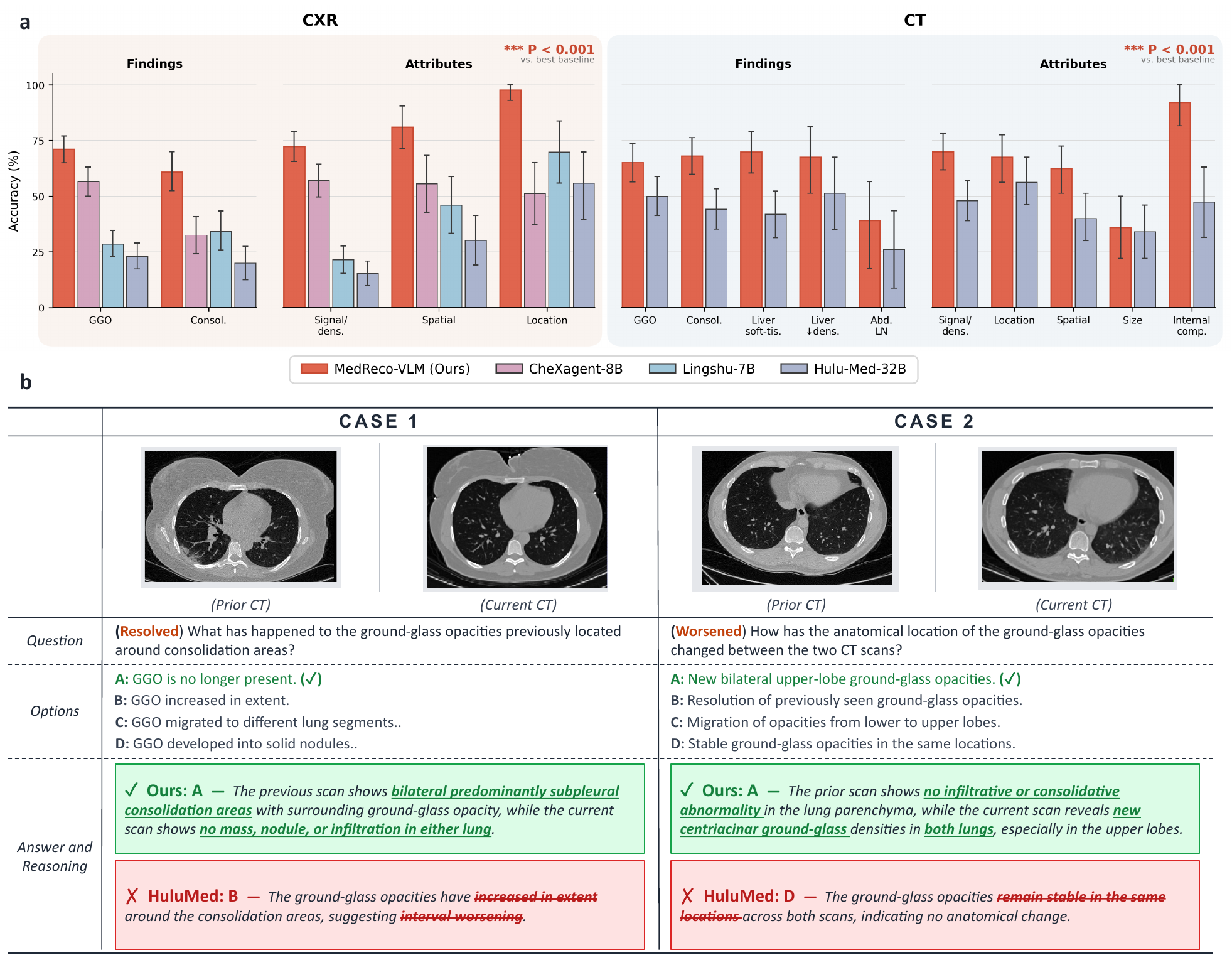}
  \vspace{4pt}
  \caption{\textbf{Generative comparative interpretation in longitudinal follow-up.}
    \textbf{a,} Accuracy on same-patient prior--current generative comparative interpretation, analysed by abnormal finding and by radiologist-prioritized follow-up attribute. Statistical significance is denoted as *\textit{P}<0.05, **\textit{P}<0.01 and ***\textit{P}<0.001. GGO, ground-glass opacity; Abd./Pelvic, abdominal/pelvic.
    \textbf{b,} Two representative longitudinal CT cases. The left case shows resolution of ground-glass opacity (GGO) previously seen around bilateral predominantly subpleural consolidation, whereas the right case shows newly developed bilateral centriacinar GGO in both lungs, especially in the upper lobes.
    }
    \vspace{-.4cm}
  \label{fig:longitudinal_follow_up}
\end{figure}

\subsection{Temporal comparison with generative comparative interpretation}

We next evaluated \genmodel~ on \textit{generative comparative interpretation}, which requires the model to compare two studies with respect to a specified clinical entity and generate an entity-specific description of similarity, difference or interval change.
Because same-patient longitudinal data are relatively limited, we first evaluated the model on a larger pool of cross-patient image pairs as a broad-scale proxy for temporal comparison. We then evaluated on same-patient longitudinal follow-up studies, in which the query consists of prior and current examinations together with a target finding or follow-up attribute, and the model must describe the interval change. 
All benchmark questions underwent multi-stage filtering, with final review by a board-certified radiologist~(Methods, Section~\ref{sec:data_construction}).
Example questions are presented in Figure~\ref{fig:dataset}.

Performance on closed-ended questions was measured using accuracy, and performance on open-ended generation was measured using BLEU~\cite{bleu}, BERTScore~\cite{zhangbertscore}, METEOR~\cite{banerjee2005meteor}, RaTEScore~\cite{zhao2024ratescore}, and RadGraph F1~\cite{delbrouck2024radgraph}, with emphasis on clinically grounded evaluation metrics. 
We compared \genmodel~ with five medical vision--language models: MedGemma-27B~\cite{sellergren2025medgemma}, RadFM-13B~\cite{RADFM}, CheXagent-8B~\cite{chen2024chexagent}, Hulu-Med-32B~\cite{jiang2025hulu}, and Lingshu-7B~\cite{xu2025lingshu}. 
Of these, only Hulu-Med-32B and RadFM-13B support 3D volumetric input. 
Results are presented across four settings: internal validation, external validation, independent public benchmarks, and same-patient longitudinal follow-up.

\textbf{Internal validation.} 
Under internal validation, \genmodel~ ranked first in all 24 evaluations across four imaging modalities and three entity levels (Figure~\ref{fig:comparison_results}a; Supplementary Table~\ref{tab:supp_comparative_indomain_compact}), with consistent gains on both closed-ended accuracy and open-ended comparative interpretation as measured by RaTEScore. 
Notably, \genmodel~ outperformed the strongest baseline by 25.5 percentage points in accuracy for CT anatomy questions and by 17.1 percentage points in RaTEScore for brain-MRI abnormality comparisons.

\textbf{Validation on external cohorts and independent public benchmarks.}
We next evaluated \genmodel~ on held-out external cohorts and independently curated public benchmarks to assess robustness to distribution shift across institutions and benchmark construction procedures (Figure~\ref{fig:comparison_results}b; Supplementary Table~\ref{tab:supp_comparative_external_public_compact}). 
On held-out external cohorts, \genmodel~ consistently outperformed baseline models, achieving 64.7\% accuracy and 60.3\% RaTEScore on chest radiographs, and 60.6\% accuracy and 57.9\% RaTEScore on CT. 
On two independent public paired-image benchmarks, Medical-Diff-VQA and MMXU, \genmodel~ achieved 87.1\% accuracy and 44.2\% RaTEScore, with a RaTEScore gain of +13.2\% relative to the second-best model.
These results indicate that the entity-aware comparative representation generalizes beyond the reporting conventions and case composition of the training institutions.

\textbf{Same-patient longitudinal follow-up studies.}
We then evaluated \genmodel~ in the clinically important setting of same-patient longitudinal follow-up, in which the key task is to determine whether a finding has progressed, resolved or remained stable. 
The test set comprised clinically prioritized findings and associated attributes that are routinely monitored in follow-up imaging, including, for example, anatomical distribution and lesion size for pulmonary consolidation (Methods, Section~\ref{sec:data_construction}). 
As shown in Figure~\ref{fig:longitudinal_follow_up}a (Supplementary Table~\ref{tab:supp_longitudinal_compact}), \genmodel~ achieved consistent gains across both modalities. 

On chest radiographs, \genmodel~ outperformed the modality-level comparator across all reported findings and follow-up attributes, with gains ranging from (+14.5\%) for pulmonary ground-glass opacity to (+46.51\%) for anatomical location.
For CT, \genmodel~ also improved over Hulu-Med-32B across all reported finding and attribute subgroups.
At the finding level, gains ranged from (+13.0\%) for abdominal/pelvic lymph nodes to (+27.9\%) for liver soft-tissue density.
At the attribute level, the largest improvement was observed for internal composition ((+44.7\%), followed by spatial distribution ((+22.5\%) percentage points) and relative signal, density or intensity ((+22.0\%).
By contrast, lesion size remained more challenging, with a smaller gain of (+2.0\%) on CT, indicating that precise quantitative assessment of temporal change remains an important direction for future work.


\textbf{Representative cases.}
Two longitudinal CT examples, each from a different patient, further illustrate the clinical relevance of correct interval-change interpretation (Figure~\ref{fig:longitudinal_follow_up}b). 
In the first case, the prior CT showed bilateral predominantly subpleural consolidation areas with surrounding ground-glass opacity, whereas the current CT showed no mass, nodule, or infiltration in either lung. 
\genmodel~ correctly identified resolution of the ground-glass opacity, whereas the baseline incorrectly reported increased extent around the consolidation areas and interpreted the interval change as worsening. 
In the second case, the prior CT showed no infiltrative or consolidative abnormality in the lung parenchyma, whereas the current CT showed new centriacinar ground-glass densities in both lungs, especially in the upper lobes. 
\genmodel~ correctly identified new bilateral upper-lobe ground-glass opacities, while the baseline incorrectly reported stable opacities in the same locations. 
These results show that \genmodel~ can generate entity-specific comparative descriptions of clinically meaningful interval change, including resolution, new abnormality development and changes in anatomical distribution.

\begin{figure}[!t]
  \centering
  \includegraphics[width=\linewidth]{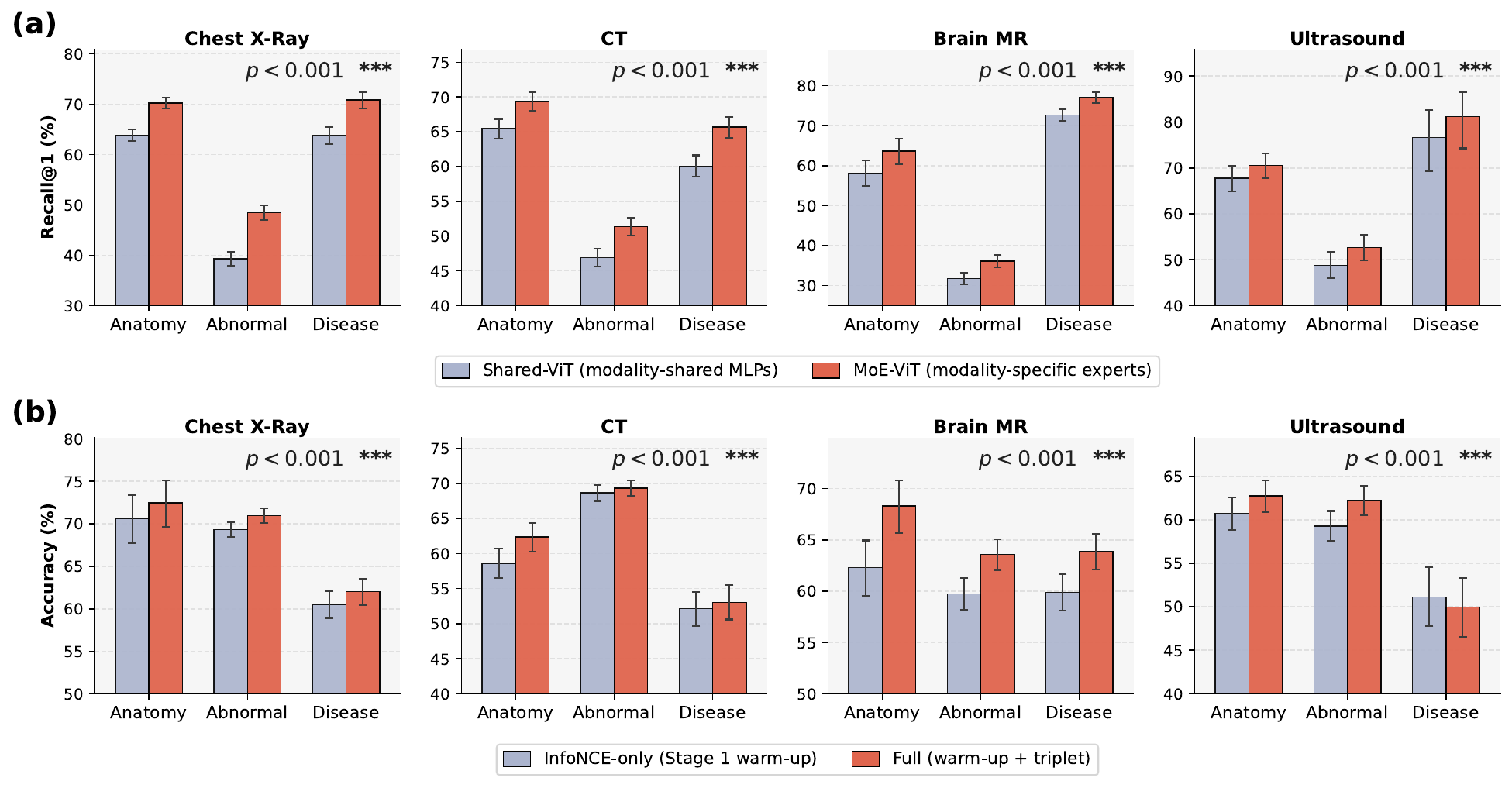}
  \caption{\textbf{Ablation studies.}
    \textbf{a,} Comparison between MoE-ViT and Shared-ViT for controllable image retrieval. Results are reported as Recall@1 (\%) on the internal validation sets across all imaging modalities and three entity granularities.
    \textbf{b,} Comparison between the Full model and the InfoNCE-only variant for generative comparative interpretation. Results are reported as closed-ended comparative-interpretation accuracy (\%) using the same modality and entity-granularity layout. Higher values indicate better performance. Error bars, 95\% CI. Top-right value, dataset-level McNemar $p$ (pooled across the three granularities). ***$p<0.001$; **$p<0.01$; *$p<0.05$.}
  \vspace{-.4cm}
  \label{fig:ablation}
\end{figure}

\subsection{Ablation studies}
\label{sec:ablation}

To identify the architectural and training components that support both \textit{controllable image retrieval} and \textit{generative comparative interpretation}, we focused on two key design elements (Section~\ref{subsec:model_architecture}): modality-specific routing in the vision encoder, which addresses heterogeneity across imaging modalities, and entity-conditioned representation learning, which organizes representations around clinically meaningful entities. 
Each ablation modified one component while keeping all other training and evaluation settings fixed, and was conducted on the internal validation sets across all imaging modalities.

\textbf{Modality-specific routing supports controllable image retrieval.}
To test whether modality-aware specialization improves controllable retrieval in \retmodel, we compared the default MoE-ViT architecture (Figure~\ref{fig:data_construction_and_model}a; Supplementary Table~\ref{tab:supp_ablation_compact} ) with a parameter-matched ViT variant in which all modalities shared the same MLP layers. MoE-ViT improved retrieval Recall@1 by a mean of 5.2 percentage points across 12 modality-entity settings and outperformed the shared-backbone ViT in 12 of 12 settings (Figure~\ref{fig:ablation}a). 
The largest gains were observed at the abnormal-finding level, suggesting that modality-specific routing is particularly beneficial for capturing fine-grained localized clinical variation across heterogeneous medical imaging modalities.


\textbf{Entity-conditioned representation learning improves generative comparative interpretation.}
To evaluate whether entity-level contrastive ranking enhances the vision encoder's capacity to capture fine-grained clinical entities, we compared \genmodel~ with a baseline variant in which the MoE-ViT is directly adapted to the LLM after initial warm-up image-report alignment~(Figure~\ref{fig:data_construction_and_model}b–c). 
\genmodel~ improved closed-ended accuracy by an average of 2.3 percentage points and outperformed the InfoNCE-only baseline in 11 of 12 settings (Figure~\ref{fig:ablation}b Supplementary Table~\ref{tab:supp_ablation_compact} ). 
These results indicate that entity-conditioned learning not only enables entity-level controllable retrieval, but also strengthens the fidelity of fine-grained visual representations required for comparative generation.
\section{Discussion}

Existing medical AI predominantly relies on a single-image paradigm, inherently limiting its capacity for comparative radiological reasoning. Consequently, tasks requiring comparative analysis, such as reference comparison for differential diagnosis and temporal comparison for longitudinal tracking, have remained underexplored. Here, we demonstrate that a unified, entity-aware framework can overcome these limitations, enabling both controllable image retrieval and generative comparative interpretation.

\textbf{Entity-conditioned representation learning.}
Conventional medical image retrieval generally relies either on global image embeddings, which do not readily isolate specific entities. By contrast, \retmodel{} learns an entity-conditioned latent space aligned to anatomical structures, abnormal findings and pathological conditions. This allows the model to distinguish among co-occurring entities within the same study and to retrieve reference cases on the basis of the target clinical concept rather than overall visual similarity. The performance on clinically confusable differential diagnoses is consistent with this interpretation, suggesting that the model captures fine-grained, entity-specific visual features instead of relying primarily on global appearance or dataset-specific correlations.

\textbf{Capturing temporal deltas in longitudinal data.}
Longitudinal assessment requires not only recognizing whether a finding is present, but also determining how it has changed between studies. In our framework, this is formulated as comparative interpretation over paired images conditioned on a target entity. The model can therefore represent interval change with respect to specific findings and generate descriptions indicating whether those findings have progressed, resolved or remained stable. The gains observed on paired-image comparison and same-patient follow-up tasks suggest that this formulation captures aspects of temporal change that are not accessible to conventional single-image models. More broadly, these results indicate that comparative supervision can extend medical vision-language modelling beyond static image description towards clinically relevant cross-study interpretation.

\textbf{Robustness across modalities and distributions.}
Trained on a large multi-institutional dataset, \retmodel{} generalized across chest radiography, CT, MRI and ultrasound, and maintained strong performance on held-out external cohorts, cross-institutional retrieval tasks and independent public benchmarks. This suggests that the model is not simply learning institution-specific visual or linguistic regularities, but is capturing comparative features that transfer across acquisition protocols, patient populations and reporting styles. Such robustness is particularly important for comparative tasks, in which clinically meaningful similarities and differences must be identified despite substantial heterogeneity in image appearance and documentation practices.

\textbf{A scalable pipeline for comparative supervision.}
Training models for comparative reasoning has historically been limited by the lack of paired, annotated data. \dataset{} addresses this by introducing an automated, scalable pipeline that transforms standard clinical reports into entity-conditioned similarity rankings and comparative VQA pairs. By leveraging this pipeline, we curated a dataset of unprecedented scale, comprising over 690,000 images from more than 160,000 patients. This methodology drastically reduces the reliance on costly manual annotations while providing the dense, structured supervision necessary to train advanced comparative vision-language models. We release the benchmark and code to the community to catalyze further research into cross-study radiological reasoning.

\textbf{Limitations and future directions.}
Although \genmodel{} performed well on interval-change interpretation, it remains limited in tasks requiring precise quantitative assessment, such as exact lesion measurement or volumetric change, which may require integration with dedicated localization or segmentation methods. 

Overall, our results suggest that comparative reasoning is a viable and important direction for medical foundation models. Rather than treating each image as an isolated input, models that compare findings across cases and across time may better reflect the way medical imaging is interpreted in practice. With broader validation, entity-conditioned comparative modelling could provide a foundation for AI systems that support reference-based reasoning, longitudinal assessment and more clinically grounded image interpretation.

\section{Method}
\label{sec:method}

\subsection{Task formulation}

Controllable medical image retrieval and generative comparative interpretation differ in their output modalities: the former produces an ordered list of relevant reference cases, whereas the latter generates natural-language descriptions of entity-specific similarities and differences. Here, we formulate the two tasks and demonstrate that they can be jointly addressed using shared visual representations, with entity-level supervisory signals.

\vspace{3pt}
\noindent \textbf{Controllable image retrieval.} 
Given a query image $I_q \in \mathbb{R}^{D \times H \times W}$ and a collection of candidate images $\mathcal{G} = \{I_1, \ldots, I_N\}$ from the same modality, the retrieval task aims to identify reference cases that match the query image on a specified clinical entity $Q$ (an anatomical structure, abnormal finding, or pathological condition). 
The retrieval model $\Phi_{\text{ret}}(\cdot)$ encodes both the query and candidate images, computes entity-conditioned similarity score between query and each candidate, with respect to $Q$, and returns a ranked list of candidates in descending order of these scores. 
\begin{align*}
    \mathcal{I}^* = \Phi_{\text{ret}}(I_q, \mathcal{G} \mid Q).
\end{align*}
When $Q$ is omitted, the model performs standard whole-image retrieval based on global visual similarity.

\vspace{3pt}
\noindent \textbf{Generative comparative interpretation.}
Given a pair of medical images $I_i, I_j \in \mathbb{R}^{D \times H \times W}$ from the same modality and an instruction $Q$ specifying a clinical entity of interest, the generative model $\Phi_{\text{gen}}(\cdot)$ produces a natural-language description $\mathcal{T}$ on the entity-specific similarities and differences between the two images:
\begin{align*}
    \mathcal{T} = \Phi_{\text{gen}}(I_i, I_j \mid Q).
\end{align*}

\vspace{3pt} \noindent \textbf{A shared vision encoder.}
Both tasks measure the \textit{entity-conditioned similarity} between two images under the same instruction $Q$: 
retrieval returns it as a ranking over candidates, generative comparative interpretation describes it in natural language. 
Therefore, building a vision encoder that captures fine-grained, entity-specific visual features across multiple imaging modalities can provide the foundational representation for both comparative reasoning tasks.

\vspace{3pt} \noindent \textbf{A shared supervisory source.}
Clinical reports already contain entity-level descriptions of the paired image. By parsing these reports into entity-specific content, we can derive complementary supervision for both tasks at scale, using the same parsed content: 
for a given entity, the \textit{agreement} between the reports provides ranking labels for controllable image retrieval, while the \textit{contrast} between them provides question-answer pairs for generative comparison interpretation.
In this way, a report-mining pipeline can produce training data for both $\Phi_{\text{gen}}(\cdot)$ and $\Phi_{\text{ret}}(\cdot)$, minimizing the need for manual annotation.

\begin{figure}[!t]
  \centering
  \includegraphics[width=\textwidth]{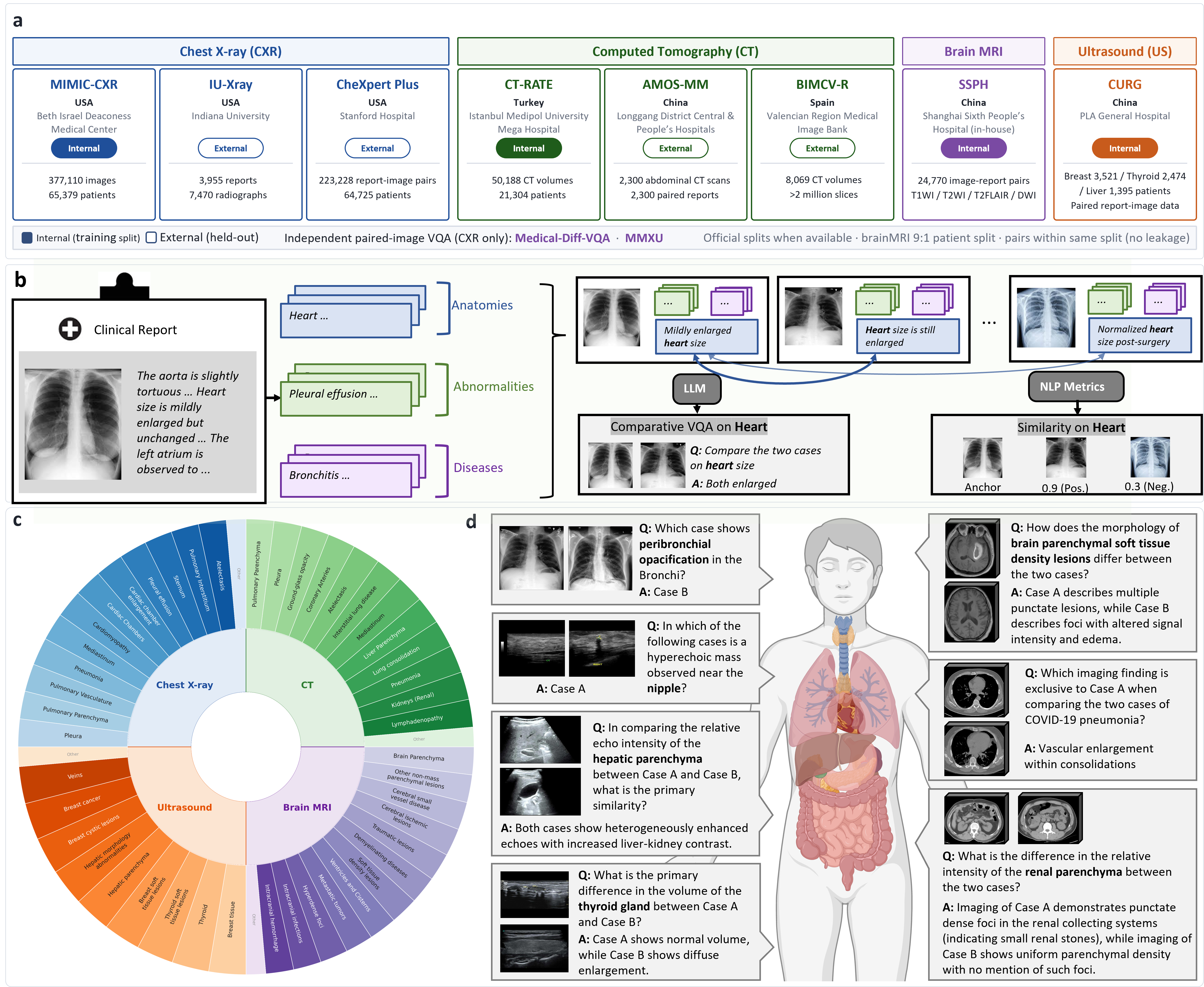}
  \vspace{2pt}
  \caption{
    \textbf{MedReCo dataset and benchmark overview.}
    \textbf{a,}~Data composition and benchmark splits across seven imaging
  modalities and eight source datasets spanning clinical centres in the
  USA, Turkey, Spain, and China. MedReCo comprises over 690{,}000 images
  from more than 160{,}000 patients; source datasets are assigned to
  internal validation or held-out external validation cohorts to evaluate
  cross-centre generalisability.
    \textbf{b,}~Comparative data construction pipeline.
Image-paired clinical reports are decomposed into structured, entity-specific descriptions at multiple granularities. For each
image pair, entity-aware similarity is quantified using three clinical NLP metrics, and comparative VQAs are generated and
carefully filtered to ensure clinical relevance and consistency. 
    \textbf{c,}~Representative entity-level coverage across modalities at
  three clinical granularities: anatomical structures, abnormal findings,
  and pathological conditions, curated per modality with radiologist review.
    \textbf{d,}~Representative comparative VQA examples across modalities.
  Each image pair is annotated with both closed-ended multiple-choice and
  open-ended questions targeting modality-specific clinical entities, with
  key clinical terms highlighted by entity type.
  }
  \vspace{-.4cm}
  \label{fig:dataset}
\end{figure}

\subsection{Data sources}
\label{sec:data_sources}

To ensure geographic, demographic and modality diversity, we assembled report-paired medical images from seven publicly available datasets and one in-house dataset, covering eight regions and seven imaging modalities. Table~\ref{tab:dataset_overview} summarises the data sources.

\noindent \textbf{Chest radiography.}
We included three chest radiography datasets. MIMIC-CXR~\cite{johnson2019mimic} comprises 377,110 images from 227,835 studies of 65,379 patients at Beth Israel Deaconess Medical Center, Boston, USA, with paired free-text radiology reports and 14 structured pathology labels. IU-Xray~\cite{IUXray} contains 3,955 chest radiographs with paired reports from Indiana University, USA. CheXpert Plus~\cite{chambon2024chexpert} provides 223,228 report--image pairs from 64,725 patients at Stanford Hospital, USA, acquired between 2002 and 2017, and includes a radiology text corpus of approximately 36 million tokens.

\noindent \textbf{Chest and abdominal computed tomography.}
We included three computed tomography datasets. CT-RATE~\cite{hamamci2026generalist} contains 50,188 non-contrast chest CT volumes from 21,304 patients at Istanbul Medipol University Mega Hospital, Turkey, with 18 multi-label abnormality annotations. AMOS-MM~\cite{ji2022amos} comprises 2,300 abdominal CT scans with bilingual Chinese--English clinical reports and 19,562 visual question-answering pairs. BIMCV-R~\cite{chen2024bimcv} contains 8,069 CT volumes, corresponding to more than 2 million slices, from the Valencian Region Medical Image Bank, Spain, covering 96 disease categories annotated by more than 20 medical professionals.

\noindent \textbf{Ultrasound.}
We used the CURG dataset~\cite{li2024ultrasound}, which contains ultrasound images with paired reports from three organ systems: breast, thyroid and liver. The dataset includes 3,521 breast, 2,474 thyroid and 1,395 liver ultrasound cases collected at PLA General Hospital, China.

\noindent \textbf{Brain magnetic resonance imaging.}
We used an in-house brain MRI dataset~\cite{LEI2025102516} comprising 24,770 image--report pairs. The dataset includes axial brain MRI examinations and covers more than ten common neurological conditions, including acute and chronic cerebral infarction, brain metastases, haemorrhage, meningioma, glioma, hydrocephalus and brain atrophy.

\subsection{Clinical entity definition}
\label{sec:entity_definition}

To reflect real-world radiological workflows, we structured the comparative analysis at three hierarchical levels of clinical granularity: anatomical structures, abnormal findings and pathological conditions. This hierarchy follows the typical process of image interpretation, progressing from anatomical localisation to the identification of imaging abnormalities and, ultimately, to diagnostic synthesis. Because clinically relevant features vary across modalities, we curated modality-specific entity sets.

\noindent \textbf{Anatomical structures.} 
Anatomical structures provide the spatial context for subsequent comparisons. We initialised the entity set using 42 anatomical categories from an established whole-body computed tomography taxonomy~\cite{zhao2026rethinking}. We then adapted this baseline to each imaging modality by retaining broadly identifiable structures and adding modality-specific anatomy, such as the nipple-areolar complex for breast ultrasound.

\noindent \textbf{Abnormal findings.} 
Abnormal findings were defined as observable and localisable deviations from normal anatomy. Starting from 404 findings in the same taxonomy~\cite{zhao2026rethinking}, we applied modality-specific filtering to retain abnormalities relevant to each imaging modality.

\noindent \textbf{Pathological conditions.} 
Pathological conditions were defined as synthesised clinical diagnoses derived from one or more imaging findings. We first extracted disease entities from radiology reports across all data sources using RaTE-NER~\cite{zhao2024ratescore}. We then used Claude 4.5 Sonnet~\cite{anthropic2025claude4} to cluster synonymous terms, such as ``pulmonary embolism'' and ``PE''. The clustered entities were manually reviewed and curated to produce a standardised disease lexicon.

This curation yielded an ontology comprising 42 anatomical structures, 69 abnormal findings and 28 pathological conditions across the seven imaging modalities in \dataset{}. A board-certified radiologist reviewed and validated all entity definitions. Modality-specific entity lists are provided in Supplementary ~\ref{sec:supp_entities}. In addition to these entity-level comparisons, we defined a \textbf{whole-image comparison} task, in which the model evaluates the complete visual content without conditioning on predefined clinical entities.

\begin{figure*}[t]
  \centering
  \includegraphics[width=\textwidth]{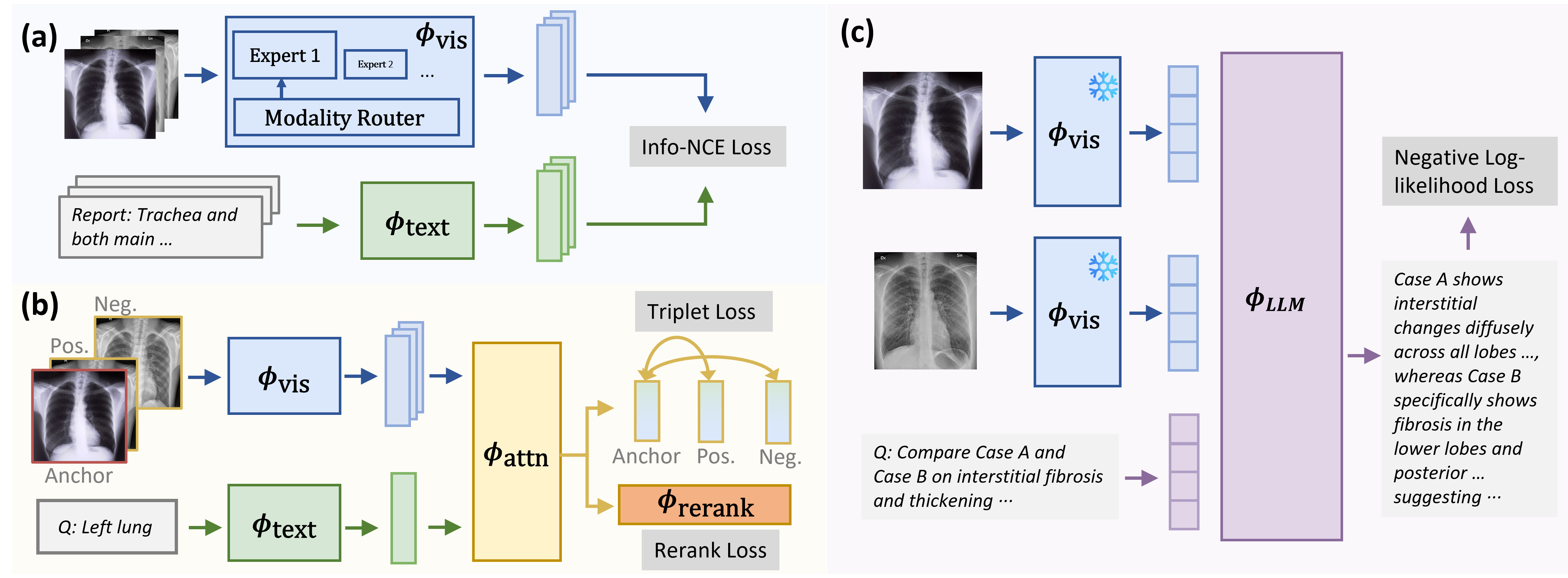}
  \vspace{2pt}
  \caption{
  \textbf{Training framework of \retmodel.}
  \textbf{(a)} Image--text representation alignment. The vision encoder is first aligned with the text encoder using image--report pairs.
  \textbf{(b)} Text-guided contrastive pre-training. The aligned vision encoder is further optimized with text-guided contrastive learning to support controllable image retrieval.
  \textbf{(c)} Instruction tuning for comparative reasoning. Finally, the pre-trained vision encoder is adapted to the LLM through instruction tuning, enabling comparative interpretation with explanatory language.
  }
  \vspace{-.4cm}
  \label{fig:data_construction_and_model}
\end{figure*}

\subsection{Data construction}
\label{sec:data_construction}

MedReCo was constructed in three stages: report decomposition, entity-level similarity quantification and comparative visual question answering (VQA) generation. The full pipeline is shown in Figure~\ref{fig:dataset}b.

\noindent\textbf{Report decomposition.}
We decomposed each radiology report into entity-specific descriptions using DeepSeek-V3.1~\cite{liu2024deepseek}. For each predefined clinical entity, the model identified whether the entity was mentioned and extracted the corresponding descriptive text. For abnormal findings and pathological conditions, we retained both positive mentions, indicating that the entity was present, and explicit negative mentions, indicating that the entity was absent. Negative mentions were retained because absence of disease or abnormality is often clinically informative in comparative interpretation. The prompts used for report decomposition are provided in Supplementary Note~\ref{sec:report_decomposition_prompts}.

\noindent\textbf{Similarity quantification.}
We estimated entity-level visual similarity using the semantic similarity of the corresponding entity-specific report descriptions as a proxy. For each image pair and clinical entity, we computed three complementary text-based similarity metrics: BioLord~\cite{remy2022biolord}, which captures biomedical semantic similarity using concept embeddings; RaTEScore~\cite{zhao2024ratescore}, which assesses factual consistency through entity-level matching; and RadGraph-XL~\cite{delbrouck2024radgraph}, which measures similarity between clinical entity--relation graphs. 
Together, these metrics provided report-derived supervision for entity-conditioned retrieval training and for benchmarking controllable image retrieval.
During both training and evaluation, an image pair was treated as a positive match only when all three entity-specific similarity metrics exceeded 0.9, whereas negative pairs used for contrastive training were defined by all three metrics being below 0.3.

\noindent\textbf{Comparative VQA generation.}
We generated a broad comparative VQA benchmark from the extracted entity-specific report descriptions. For each pair of studies from the same imaging modality, we first selected a target clinical entity from the modality-specific ontology defined in Section~\ref{sec:entity_definition} and retrieved its corresponding descriptions from both reports. We then used DeepSeek-V3.1 to compare the paired descriptions and convert clinically meaningful similarities or differences into VQA samples.

Image pairs were sampled from two sources: cross-patient pairs, which supported reference-based and differential comparison, and same-patient prior-current pairs, when available, which supported temporal comparison. For each selected image pair and entity, we generated one four-option closed-ended VQA sample with clinically plausible distractors and one corresponding open-ended VQA sample.

To ensure clinical validity and analytical difficulty, we applied a multi-stage filtering pipeline:
\vspace{-5pt}
\begin{itemize}[leftmargin=*, itemsep=2pt]
    \item \textbf{Image-dependency filtering.} 
    We prompted the language model to remove questions answerable from a single image alone, retaining only questions that required comparative reasoning across both images.
    \item \textbf{Language-prior filtering.} 
    To reduce reliance on language priors, we partitioned the VQA data and trained a text-only LLaMA-3.1-8B model~\cite{grattafiori2024llama} on each half. Each model was evaluated on the held-out split with randomised answer-option orders and queried four times per question. Questions answered correctly in at least three of four attempts were discarded.
    \item \textbf{Bias mitigation.} 
    To reduce positional bias, we randomly shuffled the multiple-choice options and swapped the presentation order of the image pairs, \(I_i \leftrightarrow I_j\).
    \item \textbf{Radiologist review.} 
    A board-certified radiologist manually reviewed the test set and removed questions that were not clinically meaningful, such as questions about compositional differences between fracture types without diagnostic relevance.
\end{itemize}

The broad comparative VQA benchmark comprised 102,052 closed-ended and 102,052 open-ended questions across dataset--entity groups. Detailed counts by dataset, modality and entity group are provided in Supplementary Table~\ref{tab:benchmark_vqa_composition}.

\noindent\textbf{Longitudinal follow-up VQAs.}
Apart from the broad comparative VQA benchmark, we constructed a dedicated longitudinal follow-up benchmark from temporally ordered same-patient prior-current studies. This benchmark was designed to assess clinically realistic interval-change reasoning rather than general cross-case comparison. It differs from the broad comparative VQA set in two ways. First, all image pairs are temporally ordered studies from the same patient. Second, the target findings and attributes are restricted to radiologist-prioritised abnormalities and follow-up attributes that are routinely monitored in longitudinal assessment, rather than being sampled from the full entity ontology. This design focuses the benchmark on clinically meaningful interval changes rather than arbitrary report differences.

For chest radiography and computed tomography, radiologists defined a checklist comprising four anatomical regions, seven key findings and follow-up attributes including size, number, margin, density and spatial distribution. For each prior-current pair, we identified candidate findings and attributes for comparison, extracted report-grounded attribute values and assigned one interval-change label: \textit{new}, \textit{resolved}, \textit{increased}, \textit{decreased}, \textit{stable}, \textit{qualitatively changed} or \textit{unable to compare}. Uncertain, clinically trivial or incomparable cases were excluded. Each retained finding-attribute change was then converted into a closed-ended VQA sample using DeepSeek-V3.1.

The resulting longitudinal benchmark comprised 930 VQA samples. Its composition is reported in Supplementary Table~\ref{tab:supp-modality}.



\subsection{Model architecture}
\label{subsec:model_architecture}
The proposed framework comprises two sequential components. First, we develop \retmodel{} for controllable medical image retrieval through text-guided image contrastive learning. This stage also serves as pre-training for the vision encoder, encouraging it to learn fine-grained visual representations of clinically meaningful entities (Section~\ref{sec:stage1}). Second, we adapt the resulting vision encoder for generative comparative interpretation by connecting it to a large language model through instruction tuning, yielding \genmodel{} (Section~\ref{sec:stage2}). The overall architecture and training procedure are illustrated in Figure~\ref{fig:data_construction_and_model}.

\subsubsection*{Stage 1: Text-guided contrastive learning for controllable retrieval}
\label{sec:stage1}
The core of \retmodel{} is a vision encoder, $\phi_{\mathrm{vis}}(\cdot)$, that learns fine-grained visual representations of clinically meaningful entities. A text encoder, $\phi_{\mathrm{text}}(\cdot)$, encodes the entity of interest, and a cross-attention module, $\phi_{\mathrm{attn}}(\cdot)$, produces entity-conditioned image embeddings for similarity estimation and retrieval. A lightweight reranker, $\phi_{\mathrm{rerank}}$, is then applied to the top-ranked candidates to refine the final ordering. We describe each component and the progressive training procedure below.

\paragraph{Vision encoder.} 
For an input image $I_k \in \mathbb{R}^{D \times H \times W}$, where $D=1$ for two-dimensional images and $D>1$ for volumetric inputs, the vision encoder produces a sequence of visual tokens:
\begin{equation}
V_k = \phi_{\mathrm{vis}}(I_k), \qquad V_k \in \mathbb{R}^{N \times d},
\end{equation}
where $N$ is the number of visual tokens and $d$ is the embedding dimension. 
We implement $\phi_{\mathrm{vis}}(\cdot)$ as a mixture-of-experts Vision Transformer variant~\cite{riquelme2021scaling}, in which images are routed to modality-specific expert multilayer perceptrons in each transformer block. This design accommodates the heterogeneity of medical imaging modalities, for example the textural appearance of ultrasound and the attenuation-based contrast of computed tomography.

To process both two-dimensional images, including radiographs and ultrasound images, and three-dimensional volumes, including CT and MRI, we decompose the transformer into two stages. The first stage comprises eight per-slice spatial self-attention layers over the in-plane dimensions. The second stage comprises four per-location self-attention layers along the depth dimension and is used only for volumetric inputs. We use a patch size of
$1 \times 20 \times 20$ for two-dimensional images and $10 \times 20 \times 20$ for three-dimensional volumes, and project all patches to a shared embedding dimension of 512.

\paragraph{Text encoder.} 
Given an entity prompt $Q$, the text encoder outputs an entity embedding:
\begin{equation}
e_Q = \phi_{\mathrm{text}}(Q), \qquad e_Q \in \mathbb{R}^{d}.
\end{equation}
We implement $\phi_{\mathrm{text}}(\cdot)$ as a BERT-style encoder initialized with pretrained BioLORD weights. The encoder takes an optional text prompt specifying the clinical entity of interest and is jointly fine-tuned during contrastive training to align entity semantics with visual representations.

\paragraph{Warm-up alignment.} 
We first align images and reports using the raw paired data and a symmetric InfoNCE loss~\cite{infonceloss} (Figure~\ref{fig:data_construction_and_model}a). This warm-up step initializes a shared multimodal embedding space from image--report pairs and stabilizes subsequent entity-level contrastive learning.

\paragraph{Entity-conditioned representation learning.}
The goal of entity-conditioned representation learning is to learn a query-dependent
view of each image, rather than assigning a single fixed representation to the
whole study. Clinically, the same image may contain multiple co-occurring entities,
and two images may be similar with respect to one entity but different with respect
to another. For example, in a chest CT containing both pulmonary artery enlargement
and hilar lymphadenopathy, the query ``pulmonary artery enlargement'' should guide
the model to focus on the pulmonary artery region and vascular caliber, whereas
the query ``hilar lymphadenopathy'' should emphasize the hilar and mediastinal
nodal regions. We therefore expect the module to learn entity-specific visual
evidence by attending to image tokens that are relevant to the queried clinical
concept while suppressing visual content that is less relevant to that entity.

To obtain entity-aware visual representation, we use a four-layer cross-attention module, $\phi_{\mathrm{attn}}(\cdot)$, that conditions the visual tokens on the entity query~(Figure~\ref{fig:data_construction_and_model}b):
\begin{equation}
f_k = \phi_{\text{attn}}(V_k, e_Q), \qquad k \in \{i, 1, \ldots, J\},
\label{eq:entity_aware_feature}
\end{equation}
where $V_k$ serves as the key and value, $e_Q$ serves as the query, $J$ denotes the number of candidate images compared against the anchor, {\em i.e.}, the query image, in the retrieval batch, and $f_k \in \mathbb{R}^{d}$ is the resulting entity-aware image feature. When $Q$ is omitted, we replace $e_Q$ with a learnable special token $\langle\mathrm{CLS}\rangle$, recovering the whole-image (entity-agnostic) setting. The entity-conditioned similarity between an anchor image $I_i$ and a candidate image $I_j$ is then computed as
\begin{equation}
S_{\text{img}}(I_i, I_j \mid Q) = f_i \cdot f_j .
\label{eq:coarse_similarity}
\end{equation}

Since $f_i$ and $f_j$ are L2-normalized, this dot product is equivalent to cosine
similarity in the entity-conditioned representation space.

\paragraph{Training objective.} Rather than regressing absolute similarity scores, we enforce the correct relative ordering of candidate images with respect to the anchor. Specifically, the supervision signal is derived from the report-based similarity, as stated in Section~\ref{sec:data_construction}:
\begin{equation}
\mathcal{I}(S_{\text{img}}(I_i, I_j \mid Q)) = \mathcal{I}(S_{\text{rpt}}(R_i, R_j \mid Q)), \quad j \in \{1, \ldots, J\}
\end{equation}
where $\mathcal{I}$ is the ranking function, and $S_{\text{rpt}}$ denotes the semantic similarity between entity-specific contents in reports $R_i$ and $R_j$, quantified by the averaged ranking from BioLord, RaTEScore, and RadGraph-XL. This objective is implemented with a triplet loss~\cite{Triplet_loss}:
\begin{equation}
\mathcal{L}_{\text{triplet}} = \sum_{(i,p,n)} \max\left(0, \; S_{\text{img}}(I_i, I_n \mid Q) - S_{\text{img}}(I_i, I_p \mid Q) + \alpha\right)
\end{equation}
where $(I_i, I_p, I_n)$ denotes an anchor--positive--negative triplet, defined in Section~\ref{sec:data_construction}, and $\alpha$ is the margin, set to 0.2. 
To ensure robust fine-grained representation learning, we employ a hard negative mining strategy within each mini-batch.

\paragraph{Coarse-to-fine re-ranking.}
The coarse retriever ranks the gallery efficiently using the dot product between the entity-aware image features in Eq.~\ref{eq:coarse_similarity}. However, representing each image with a single vector may discard localized visual evidence that is important for clinically confusable entities. We therefore apply a lightweight point-wise reranker only to the top candidates returned by coarse retrieval, enabling fine-grained token-level matching with limited computational cost.

For each image $I_k$ in a query--candidate pair, where $k \in \{i,j\}$, we reuse the final-layer cross-attention map from Eq.~\ref{eq:entity_aware_feature} to identify the visual patches most relevant to the queried entity $Q$. Specifically, let $A_k \in \mathbb{R}^{N}$ denote the final-layer cross-attention weights
from the entity query to the $N$ visual tokens, averaged over attention heads.
We select the indices of the top-$K$ attended visual tokens as
\begin{equation}
\mathcal{P}_k = \operatorname{TopK}(A_k), \qquad |\mathcal{P}_k|=K.
\end{equation}
The compact entity-conditioned token sequence is then constructed as
\begin{equation}
\tilde{V}_k =
\left[
f_k;\,
V_{k,\mathcal{P}_k}
\right],
\qquad
\tilde{V}_k \in \mathbb{R}^{(K+1)\times d},
\label{eq:compact_entity_tokens}
\end{equation}
where $V_{k,\,\mathcal{P}_k} \in \mathbb{R}^{K \times d}$ denotes the
subset of visual tokens in~$V_k$ indexed by~$\mathcal{P}_k$.  We set $K{=}64$ in all experiments.

The reranker takes the compact token sequences from the query and candidate images, along with a learnable $\langle\mathrm{CLS}\rangle$ token, as input:
\begin{equation}
Z_{i,j} =
\left[
\langle\mathrm{CLS}\rangle;\,
\tilde{V}_i;\,
\tilde{V}_j
\right].
\end{equation}

The sequence $Z_{i,j}$ is processed by a shallow modality-aware MoE self-attention encoder, denoted as $\phi_{\mathrm{rerank}}$, where self-attention models fine-grained interactions between the query and
candidate visual tokens, and modality-specific experts adapt the feed-forward layers to different imaging modalities. Let
\begin{equation}
H_{i,j}=\phi_{\mathrm{rerank}}(Z_{i,j}),
\end{equation}
where $H_{i,j}$ denotes the output token sequence of the reranker encoder.
A linear classification head $h_{\mathrm{cls}}$ is then applied to the
output $\langle\mathrm{CLS}\rangle$ token to produce the refined relevance
score,
\begin{equation}
  s^{\mathrm{rerank}}_{i,j}
  =
  \sigma\!\left(
  h_{\mathrm{cls}}\!\left(H_{i,j}^{\mathrm{CLS}}\right)
  \right),
  \label{eq:rerank_score}
\end{equation}
where $H_{i,j}^{\mathrm{CLS}}$ denotes the output representation
corresponding to the $\langle\mathrm{CLS}\rangle$ token, and
$\sigma(\cdot)$ denotes the sigmoid function.

\subsubsection*{Stage 2: Instruction tuning for comparative interpretation}
\label{sec:stage2}

As the pre-trained vision encoder already captures fine-grained visual representations of clinically meaningful entities, we leverage these representations to build $\Phi_{\text{gen}}(\cdot)$ for generating detailed textual comparisons.

\paragraph{Architecture.} 
As shown in Figure~\ref{fig:data_construction_and_model}c, we bridge $\phi_{\text{vis}}(\cdot)$ to a large language model $\phi_{\text{LLM}}(\cdot)$ through a projection layer. 
The projection layer consists of a two-layer MLP with a GELU activation function, mapping the visual features into the hidden space of LLM. The comparative interpretation is formulated as:
\begin{equation}
T=\phi_{\text{LLM}}
\Big(Q,\;\Pi(\phi_{\text{vis}}(I_i)),\;\Pi(\phi_{\text{vis}}(I_j))\Big),
\end{equation}
where $\Pi(\cdot)$ denotes the projection module, $T$ and $Q$ are the generated comparison and the instruction from our VQA pairs, respectively.
We adopt Qwen2.5-7B-Instruct~\cite{qwen2.5} as $\phi_{\text{LLM}}(\cdot)$, 
with all parameters trainable during instruction tuning.

\paragraph{Training objective.} 
The model is trained with standard negative log-likelihood loss for auto-regressive next-token prediction:
\begin{equation}
\mathcal{L}_{\text{NLL}}
=-\sum_{t=1}^{|T|}\log P_{\phi_{\text{LLM}}}
\!\left(T_t \mid T_{<t}, Q, \Pi(\phi_{\text{vis}}(I_i)), \Pi(\phi_{\text{vis}}(I_j))\right).
\end{equation}
where $T_t$ denotes the $t$-th output token and $T_{<t}$ denotes the previously generated tokens. 
During Stage 2, the vision encoder $\phi_{\text{vis}}$ is frozen to preserve the entity-aware representations learned in Stage 1, while the projection module $\Pi(\cdot)$ and the LLM $\phi_{\text{LLM}}$ are optimized for comparative generation.

\subsection*{Training details}
Stage 1 is optimized with AdamW using linear warm-up and cosine decay, with per-GPU batch sizes of 14 for 3D inputs and 50 for 2D inputs, for 10 epochs. Stage 2 also uses AdamW, with learning rates of $2\times10^{-5}$ for the LLM and $1\times10^{-4}$ for the projection module, a global batch size of 64, and one epoch of instruction tuning. All experiments are conducted on $8\times$A100 (80\,GB) GPUs. Full hyperparameters are provided in Supplementary Table~\ref{tab:training_hyperparameters}.

\section{Data Availability}
All public datasets used in this study are available from their original data providers. 
Official access links, dataset roles, and access requirements for the public datasets and independent paired-image VQA benchmarks are provided in Supplementary Table~\ref{tab:dataset_overview}. 
These datasets should be accessed and used in accordance with the licenses, data-use agreements, and approval requirements of the original providers.
The in-house SSPH brain MRI dataset is not publicly available because it contains institutionally collected clinical imaging data subject to patient-privacy and data-governance restrictions. 

Derived annotations, benchmark splits, and benchmark-construction scripts will be released with the project repository upon publication, where permitted by the licenses and data-use agreements of the original datasets.

\section{Code Availability}
The code and benchmark will be released at https://github.com/time-seed/MedReCo.

\section{Acknowledgements}
This work is supported by the Innovative Drug Research and Development National Science and Technology Major Project (No. 2025ZD1803101), the Scientific Research Innovation Capability Support Project for Young Faculty (ZYGXQNJSKYCXNLZCXM-I22), the National Natural Science Foundation of China (No. 24Z031503678), and the Science and Technology Innovation Action Plan of Shanghai Municipality (No. 24QA2703800).

\bibliographystyle{sn-mathphys} 
\bibliography{main} 
\clearpage

\appendix
\clearpage
\setcounter{page}{1}
\setcounter{section}{0}
\setcounter{table}{0}
\setcounter{figure}{0}
\renewcommand{\thesection}{S\arabic{section}}
\renewcommand{\thesubsection}{S\arabic{section}.\arabic{subsection}}
\renewcommand{\thetable}{S\arabic{table}}
\renewcommand{\thefigure}{S\arabic{figure}}

\hypersetup{linkcolor=black,citecolor=black,urlcolor=black}
\setlength{\parskip}{2pt}
\setlength{\parindent}{0pt}

\renewcommand{\arraystretch}{1.16}
\setlength{\extrarowheight}{0.8pt}

\newcommand{\suppTableTopSpace}{\par\vspace{10pt}}
\newcommand{\suppTableBottomSpace}{\par\vspace{10pt}}

\makeatletter
\setlength{\@fpsep}{10pt plus 2pt minus 1pt}
\setlength{\@dblfpsep}{10pt plus 2pt minus 1pt}
\makeatother

\makeatletter
\setlength{\@fptop}{0pt}
\setlength{\@fpsep}{8pt plus 2pt minus 2pt}
\setlength{\@fpbot}{0pt plus 1fil}
\setlength{\@dblfptop}{0pt}
\setlength{\@dblfpsep}{8pt plus 2pt minus 2pt}
\setlength{\@dblfpbot}{0pt plus 1fil}
\makeatother

\providecommand{\NA}{\textemdash}
\providecommand{\Internal}{\textsc{Internal}}
\providecommand{\External}{\textsc{External}}
\providecommand{\Benchmark}{\textsc{Benchmark}}
\providecommand{\ourretrieval}{MedReCo}
\providecommand{\ours}{MedReCo-VLM}
\providecommand{\datasetgroup}[1]{%
  \addlinespace[2pt]
  \multicolumn{8}{@{}l}{\textit{\bfseries #1}}\\[-1pt]%
}

\providecommand{\tablefootnote}[1]{%
  \suppTableBottomSpace
  \begin{minipage}{\linewidth}
  \footnotesize\emph{Note.}~#1
  \end{minipage}%
  \par\vspace{2pt}%
}
\lstdefinestyle{suppPromptListing}{%
  basicstyle=\ttfamily\fontsize{7.2}{8.3}\selectfont,
  breaklines=true,
  breakatwhitespace=false,
  columns=fullflexible,
  keepspaces=true,
  showstringspaces=false,
  frame=single,
  framerule=0.35pt,
  framesep=4pt,
  aboveskip=6pt,
  belowskip=6pt,
  captionpos=t
}

\lstnewenvironment{promptbox}[1]{%
  \lstset{style=suppPromptListing,title={#1}}%
}{}

\setlist[description]{%
  leftmargin=2.8cm,
  labelwidth=2.55cm,
  labelsep=0.25cm,
  itemsep=3pt,
  parsep=0pt,
  topsep=3pt,
  font=\normalfont\bfseries
}

\section{Supplementary Dataset and Construction Details}
\label{sec:supp_dataset_construction}

\subsection{Ethics Approval.} 
The collection and use of the in-house brain MRI dataset was approved by the Institutional Review Board of Shanghai Sixth People's Hospital (IRB code: 2023-KY-082(K)). 
Informed consent was waived owing to the retrospective and non-interventional nature of the study and the use of fully de-identified data. 
All other datasets used in this study are publicly available and were used in accordance with their respective data use agreements. 
All data were de-identified before being transferred to study investigators. 

\subsection{Dataset sources and roles}
\label{sec:supp_datasets}

MedReCo draws on eight image--report datasets spanning four imaging modality groups: chest X-ray (CXR), computed tomography (CT), brain magnetic resonance imaging (BRAIN), and ultrasound (US). Two independent paired-image VQA benchmarks are used solely for evaluation. Table~\ref{tab:dataset_overview} summarizes the source, modality, scale, access route, and role of each dataset.

\begin{table*}[!htbp]
\centering
\scriptsize
\setlength{\tabcolsep}{3.0pt}
\caption{Dataset sources used in MedReCo. \textsuperscript{\textdagger}In-house datasets are not publicly available.}
\label{tab:dataset_overview}
\suppTableTopSpace
\begin{tabularx}{\textwidth}{>{\raggedright\arraybackslash}p{2.1cm}>{\raggedright\arraybackslash}p{3.3cm}>{\raggedright\arraybackslash}p{1.25cm}>{\raggedleft\arraybackslash}p{1.2cm}>{\raggedleft\arraybackslash}p{1.35cm}>{\raggedright\arraybackslash}p{2.55cm}>{\raggedright\arraybackslash}X>{\centering\arraybackslash}p{1.35cm}}
\toprule
\textbf{Dataset} & \textbf{Country / centre} & \textbf{Modality} & \textbf{\#Patients} & \textbf{\#Studies} & \textbf{\#Images / volumes} & \textbf{Access} & \textbf{Role} \\
\midrule
\datasetgroup{Chest X-ray (CXR)}
MIMIC-CXR & USA / Beth Israel Deaconess Medical Center & CXR & 65,379 & 227,835 & 377,110 images & \href{https://physionet.org/content/mimic-cxr/2.1.0/}{PhysioNet} & \Internal{} \\
IU-Xray & USA / Indiana University & CXR & 3,955 & 3,955 & 7,470 radiographs & \href{https://www.kaggle.com/datasets/raddar/chest-xrays-indiana-university}{Kaggle} & \External{} \\
CheXpert Plus & USA / Stanford Hospital & CXR & 64,725 & 187,711 & 223,228 images & \href{https://stanfordaimi.azurewebsites.net/datasets/5158c524-d3ab-4e02-96e9-6ee9efc110a1}{Stanford AIMI} & \External{} \\
\datasetgroup{Computed tomography (CT)}
CT-RATE & Turkey / Istanbul Medipol Univ. Mega Hospital & Chest CT & 21,304 & 25,692 & 50,188 volumes & \href{https://huggingface.co/datasets/ibrahimhamamci/CT-RATE}{Hugging Face} & \Internal{} \\
AMOS-MM & China / Longgang District Central \& People's Hospital & Abdominal CT & 2,300 & 2,300 & 2,300 scans & \href{https://zenodo.org/records/10992155}{Zenodo} & \Internal{} \\
BIMCV-R & Spain / Valencian Region Medical Image Bank & Whole-body CT & 8,069 & 8,069 & 8,069 vol. ($>$2M slices) & \href{https://huggingface.co/datasets/cyd0806/BIMCV-R}{Hugging Face} & \External{} \\
\datasetgroup{Brain MRI (BRAIN)}
SSPH\textsuperscript{\textdagger} & China / Shanghai Sixth People's Hospital & Brain MRI & 24,770 & 24,770 & 24,770 pairs (4 seq.) & Not publicly available & \Internal{} \\
\datasetgroup{Ultrasound (US)}
CURG & China / PLA General Hospital & US & 7,390 & 7,390 & Paired report--image & \href{https://github.com/LijunRio/Ultrasound-Report-Generation}{GitHub} & \Internal{} \\
\midrule
\datasetgroup{Independent paired-image VQA benchmarks (CXR)}
Medical-Diff-VQA & Derived from MIMIC-CXR & CXR & \NA & 164,324 pairs & 164,324 QA (diff. subset) & \href{https://physionet.org/content/medical-diff-vqa/1.0.1/}{PhysioNet} & \Benchmark{} \\
MMXU & Derived from MIMIC-CXR & CXR & 1,201 & 2,469 & 3K test + 118K dev QA & \href{https://github.com/LinjieMu/MMXU}{GitHub} & \Benchmark{} \\
\bottomrule
\end{tabularx}
\tablefootnote{SSPH contains brain MRI--report pairs across four axial sequences: T1WI, T2WI, T2FLAIR, and DWI. CURG contains organ-specific ultrasound subsets for breast (3,521), thyroid (2,474), and liver (1,395) imaging.}
\end{table*}

SSPH contains 24,770 brain MRI--report pairs across four sequences (T1WI, T2WI, T2FLAIR, and DWI), covering 13 disease categories with a 9:1 patient-level train/test split. CURG comprises three organ-specific ultrasound sub-datasets: breast (3,521 patients), thyroid (2,474 patients), and liver (1,395 patients), each with paired Chinese-language diagnostic reports. Medical-Diff-VQA provides 700,703 QA pairs in seven categories; we use the \textit{difference} subset (164,324 QA; train/validation/test = 131,563/16,372/16,389). MMXU evaluates region-specific disease changes via multiple-choice questions.

\subsection{Split construction and leakage prevention}
\label{sec:supp_split_leakage}

All data splits are constructed at the patient level whenever patient identifiers are available. Derived samples---including image pairs, anchor--positive--negative triplets, closed-ended questions, and open-ended questions---are generated only after split assignment. This prevents train--test contamination across both retrieval and generative-comparison tasks.

\begin{table*}[!htbp]
\centering
\small
\caption{Leakage-control rules used during construction of derived comparative samples.}
\label{tab:leakage_control}
\suppTableTopSpace
\begin{tabularx}{\textwidth}{>{\raggedright\arraybackslash}p{3.1cm}>{\raggedright\arraybackslash}X>{\raggedright\arraybackslash}X>{\raggedright\arraybackslash}p{3.0cm}}
\toprule
\textbf{Derived sample type} & \textbf{Construction rule} & \textbf{Leakage-control rule} & \textbf{Used for} \\
\midrule
Entity-level report content & Extracted independently from each report for predefined entities. & Extraction is performed after dataset access control; no cross-split context is used. & Retrieval and VQA supervision \\
Cross-patient image pair & Two cases from the same modality group are paired to compare a target entity. & Both cases must come from the same split. & Broad comparative VQA \\
Same-patient longitudinal pair & Prior and current studies are ordered temporally within the same patient. & The patient cannot appear in more than one split. & Longitudinal follow-up VQA \\
Retrieval triplet & Anchor, positive, and negative are selected according to entity-level report similarity. & Anchor, positive, and negative must come from the same split during training/evaluation construction. & Stage-1 contrastive ranking \\
External retrieval gallery & Query and gallery are formed from held-out centres. & External cohorts are excluded from training and checkpoint selection. & External and cross-centre validation \\
\bottomrule
\end{tabularx}
\end{table*}

\subsection{Report-mining and comparative-supervision pipeline}
\label{sec:supp_pipeline}

MedReCo is constructed from paired images and radiology reports through a shared report-mining pipeline. The same parsed entity-level report content supports both reference comparison and temporal comparison: entity-level agreement between reports provides ranking supervision for controllable retrieval, whereas entity-level contrast between reports provides instruction--response supervision for comparative interpretation.

\begin{table*}[!htbp]
\centering
\small
\caption{Overview of the report-mining and comparative-supervision pipeline.}
\label{tab:pipeline_overview}
\suppTableTopSpace
\begin{tabularx}{\textwidth}{>{\centering\arraybackslash}p{0.8cm}>{\raggedright\arraybackslash}p{3.0cm}>{\raggedright\arraybackslash}X>{\raggedright\arraybackslash}X}
\toprule
\textbf{Step} & \textbf{Module} & \textbf{Input} & \textbf{Output} \\
\midrule
1 & Entity-specific report decomposition & Radiology report and a predefined entity from the modality-specific vocabulary. & Entity-specific text span or null output; explicit positive and negative statements are retained. \\
2 & Mention-status assignment & Extracted text span for anatomy, abnormality, or disease. & Mention status: positive, explicitly negative, uncertain, or not mentioned. \\
3 & Disease-term normalization & Disease mentions extracted from reports. & Standardized disease terms after synonym clustering and manual curation. \\
4 & Entity-level similarity quantification & Entity-specific report content from two studies. & Report-derived similarity scores used to supervise entity-conditioned visual similarity. \\
5 & Retrieval triplet construction & Anchor study, candidate studies, entity condition, and report-derived similarity. & Anchor--positive--negative triplets for text-guided contrastive ranking. \\
6 & Comparative VQA generation & Two reports, target anatomy, target abnormality or disease, and entity-specific descriptions. & Multiple-choice comparative questions with answer and rationale. \\
7 & Open-ended QA conversion & Multiple-choice question, options, answer, and rationale. & Short-answer comparative question--answer pair. \\
8 & VQA filtering and review & Generated closed-ended and open-ended questions. & Final VQA samples requiring genuine paired-image reasoning. \\
\bottomrule
\end{tabularx}
\end{table*}

\paragraph{Report decomposition.} For each report, we extract entity-specific content at three levels: anatomy, abnormal finding, and disease. Anatomy prompts extract all report content related to a specific structure. Abnormality prompts extract the presence, location, size, morphology, or characteristics of a target abnormality within a target anatomy. Disease prompts extract content directly related to a target disease or condition. Explicitly negated statements are retained because the absence of an entity can be clinically meaningful when comparing studies.

\paragraph{Comparative VQA generation.} For each pair of reports, we generate questions that compare Case~A and Case~B with respect to a specific target abnormality or entity. The questions focus on dimensions such as presence/absence, location, size, morphology, margins, internal characteristics, associated findings, and potential etiology.

\paragraph{VQA filtering and review.} Generated questions are filtered to remove items that do not require two images, are answerable from language priors alone, contain biased or duplicate answer patterns, or lack clinical specificity. The filtering stages are summarized in Table~\ref{tab:vqa_filtering}.

\begin{table*}[!htbp]
\centering
\small
\caption{Filtering stages for comparative VQA construction.}
\label{tab:vqa_filtering}
\suppTableTopSpace
\begin{tabularx}{\textwidth}{>{\raggedright\arraybackslash}p{3.0cm}>{\raggedright\arraybackslash}X>{\raggedright\arraybackslash}X>{\raggedright\arraybackslash}p{2.5cm}}
\toprule
\textbf{Filtering stage} & \textbf{Retain} & \textbf{Remove} & \textbf{Output} \\
\midrule
Image-dependency filter & Questions that strictly require examining both Case~A and Case~B. & Single-image identification, isolated anatomy description, or non-comparative questions. & Paired-image-dependent VQA \\
Language-prior filter & Questions whose answer cannot be inferred from the wording alone. & Questions with deterministic linguistic cues, trivial option structure, or obvious answer priors. & Visually grounded VQA \\
Bias and duplicate filter & Balanced answer distribution and non-duplicated question forms. & Repeated stems, near-duplicate options, and systematic answer-position bias. & De-duplicated VQA \\
Clinical review & Clinically meaningful, factual, and unambiguous comparisons. & Clinically irrelevant, ambiguous, contradictory, or insufficiently grounded samples. & Final comparative VQA \\
\bottomrule
\end{tabularx}
\end{table*}

\subsection{Benchmark composition}
\label{sec:supp_benchmark_composition}

We further summarize the sample-level composition of the benchmarks used for entity-conditioned retrieval and generative comparative VQA. For retrieval, counts are reported as retained query--entity sample rows after filtering. Internal samples are split into training and held-out test sets, whereas external samples are drawn only from held-out datasets. For VQA, counts are reported by source dataset and entity level.

\begin{table*}[!htbp]
\centering
\scriptsize
\setlength{\tabcolsep}{3.5pt}
\caption{Benchmark composition for entity-conditioned retrieval. Counts indicate retained query--entity sample rows after filtering.}
\label{tab:benchmark_retrieval_composition}
\suppTableTopSpace
\begin{tabularx}{\textwidth}{>{\raggedright\arraybackslash}p{1.35cm}>{\raggedright\arraybackslash}p{1.85cm}>{\raggedleft\arraybackslash}p{1.8cm}>{\raggedleft\arraybackslash}p{1.7cm}>{\raggedleft\arraybackslash}p{1.7cm}>{\raggedright\arraybackslash}X}
\toprule
\textbf{Modality group} & \textbf{Entity level} & \textbf{Internal train} & \textbf{Internal test} & \textbf{External test} & \textbf{External source(s)} \\
\midrule
BRAIN & Abnormality & 37,296 & 4,242 & \NA & \NA \\
BRAIN & Anatomy     & 20,540 & 2,281 & \NA & \NA \\
BRAIN & Disease     & 63,223 & 7,090 & \NA & \NA \\
\addlinespace[2pt]
CT & Abnormality & 97,102  & 6,376 & 3,770 & BIMCV-R \\
CT & Anatomy     & 115,547 & 7,511 & 5,174 & BIMCV-R \\
CT & Disease     & 77,752  & 6,026 & 4,120 & BIMCV-R \\
\addlinespace[2pt]
CXR & Abnormality & 294,034 & 5,672  & 5,846 & CheXpert Plus (2,094); IU-Xray (3,752) \\
CXR & Anatomy     & 557,330 & 10,784 & 9,240 & CheXpert Plus (3,736); IU-Xray (5,504) \\
CXR & Disease     & 122,026 & 5,070  & 4,988 & CheXpert Plus (1,624); IU-Xray (3,364) \\
\addlinespace[2pt]
US & Abnormality & 13,290 & 1,426 & \NA & \NA \\
US & Anatomy     & 15,242 & 1,640 & \NA & \NA \\
US & Disease     & 2,062  & 262   & \NA & \NA \\
\midrule
\textbf{Total} & \textbf{All} & \textbf{1,293,418} & \textbf{58,380} & \textbf{33,138} & CheXpert Plus, IU-Xray, and BIMCV-R \\
\bottomrule
\end{tabularx}
\tablefootnote{Internal retrieval samples are grouped by modality family. External test samples are from held-out datasets only: CheXpert Plus and IU-Xray for CXR, and BIMCV-R for CT. No external BRAIN or US retrieval cohort was included in the current benchmark.}
\end{table*}

\begin{table*}[!htbp]
\centering
\scriptsize
\setlength{\tabcolsep}{5pt}
\caption{Test-set composition for generative comparative VQA. Counts indicate the number of VQA samples by source dataset and entity level.}
\label{tab:benchmark_vqa_composition}
\suppTableTopSpace
\begin{tabularx}{\textwidth}{>{\raggedright\arraybackslash}p{2.45cm}>{\raggedleft\arraybackslash}p{2.0cm}>{\raggedleft\arraybackslash}p{2.0cm}>{\raggedleft\arraybackslash}p{2.0cm}>{\raggedleft\arraybackslash}p{2.1cm}>{\raggedleft\arraybackslash}p{2.0cm}}
\toprule
\textbf{Dataset} & \textbf{Abnormality} & \textbf{Anatomy} & \textbf{Disease} & \textbf{Per-version total} & \textbf{Overall total} \\
\midrule
AMOS-MM        & 7,212  & 8,794  & 14,430 & 15,218 & 30,436 \\
BIMCV-R        & 4,798  & 972    & 600    & 3,185  & 6,370 \\
Brain MRI      & 15,606 & 4,996  & 11,684 & 16,143 & 32,286 \\
CT-RATE        & 24,932 & 8,496  & 12,726 & 23,077 & 46,154 \\
CheXpert Plus  & 1,746  & 3,424  & 1,820  & 3,495  & 6,990 \\
MIMIC-CXR      & 21,614 & 1,982  & 7,522  & 15,559 & 31,118 \\
CURG / US      & 36,798 & 10,622 & 3,330  & 25,375 & 50,750 \\
\midrule
\textbf{Total} & \textbf{112,706} & \textbf{39,286} & \textbf{52,112} & \textbf{102,052} & \textbf{204,104} \\
\bottomrule
\end{tabularx}
\tablefootnote{The source statistics contain two matched VQA test variants, denoted as ``1113'' and ``free''. The entity-level columns and the overall total sum both variants; the per-version total reports the size of either variant. Independent public CXR VQA benchmarks are reported separately in Table~\ref{tab:dataset_overview}: Medical-Diff-VQA uses the difference-subset test split with 16,389 QA, and MMXU contains 3K test QA.}
\end{table*}

\begin{table}[htbp]
\centering
\footnotesize
\begin{threeparttable}
\caption{Composition of the longitudinal follow-up VQA benchmark by modality and clinical target.}
\label{tab:supp-modality}
\begin{tabular}{llllrr}
\toprule
Modality & Source cohort & Anatomical group & Finding group & Samples & Percentage \\
\midrule
CXR & CheXpert Plus & Pulmonary parenchyma & Ground-glass opacity & 343 & 36.88\% \\
CXR & CheXpert Plus & Pulmonary parenchyma & Lung consolidation & 195 & 20.97\% \\
\textbf{CXR subtotal} & --- & --- & --- & \textbf{538} & \textbf{57.85\%} \\
\midrule
CT & CT-RATE & Pulmonary parenchyma & Ground-glass opacity & 126 & 13.55\% \\
CT & CT-RATE & Pulmonary parenchyma & Lung consolidation & 122 & 13.12\% \\
CT & CT-RATE & Liver / hepatic parenchyma & Soft-tissue density lesion & 81 & 8.71\% \\
CT & CT-RATE & Liver / hepatic parenchyma & Decreased hepatic parenchymal density & 36 & 3.87\% \\
CT & CT-RATE & Abdominopelvic cavity & Peritoneal findings & 27 & 2.90\% \\
\textbf{CT subtotal} & --- & --- & --- & \textbf{392} & \textbf{42.15\%} \\
\midrule
\textbf{Total} & --- & --- & --- & \textbf{930} & \textbf{100.00\%} \\
\bottomrule
\end{tabular}
\begin{tablenotes}[flushleft]
\footnotesize
\item Percentages were calculated over the final filtered longitudinal follow-up VQA benchmark of 930 samples. Samples labelled as unable to compare or uncompare were excluded before tabulation. Categories with fewer than five samples after pooling CXR and CT were merged into broader clinically related categories and were not reported as separate strata.
\end{tablenotes}
\end{threeparttable}
\end{table}

\begin{table}[htbp]
\centering
\footnotesize
\begin{threeparttable}
\caption{Composition of the longitudinal follow-up VQA benchmark by follow-up attribute and interval-change label.}
\label{tab:supp-attribute}
\begin{tabular}{llrrrr}
\toprule
Category type & Category & CXR & CT & Total & Percentage \\
\midrule
Follow-up attribute & Relative signal / density / intensity & 288 & 122 & 410 & 44.09\% \\
Follow-up attribute & Spatial distribution & 121 & 80 & 201 & 21.61\% \\
Follow-up attribute & Anatomical location & 72 & 80 & 152 & 16.34\% \\
Follow-up attribute & Finding presence / status & 47 & 2 & 49 & 5.27\% \\
Follow-up attribute & Lesion size & 0 & 48 & 48 & 5.16\% \\
Follow-up attribute & Internal composition & 0 & 36 & 36 & 3.87\% \\
Follow-up attribute & Margins & 0 & 13 & 13 & 1.40\% \\
Follow-up attribute & Number of lesions & 0 & 11 & 11 & 1.18\% \\
Follow-up attribute & Other / rare attributes & 10 & 0 & 10 & 1.08\% \\
\textbf{Follow-up attribute subtotal} & --- & \textbf{538} & \textbf{392} & \textbf{930} & \textbf{100.00\%} \\
\midrule
Interval-change label & Resolved & 177 & 97 & 274 & 29.46\% \\
Interval-change label & Stable & 150 & 96 & 246 & 26.45\% \\
Interval-change label & New & 97 & 117 & 214 & 23.01\% \\
Interval-change label & Qualitatively changed & 40 & 53 & 93 & 10.00\% \\
Interval-change label & Increased & 40 & 24 & 64 & 6.88\% \\
Interval-change label & Decreased & 34 & 5 & 39 & 4.19\% \\
\textbf{Interval-change subtotal} & --- & \textbf{538} & \textbf{392} & \textbf{930} & \textbf{100.00\%} \\
\bottomrule
\end{tabular}
\begin{tablenotes}[flushleft]
\footnotesize
\item Percentages were calculated over the final filtered longitudinal follow-up VQA benchmark of 930 samples. Samples labelled as unable to compare or uncompare were excluded before tabulation. Categories with fewer than five samples after pooling CXR and CT were merged into broader clinically related categories and were not reported as separate strata.
\end{tablenotes}
\end{threeparttable}
\end{table}

\section{Supplementary Clinical Entity Groups}
\label{sec:supp_entities}

\begin{table*}[!htbp]
\centering
\small
\caption{Summary of clinical entity groups by modality family.}
\label{tab:entity_group_summary}
\suppTableTopSpace
\begin{tabularx}{\textwidth}{>{\raggedright\arraybackslash}p{1.8cm}>{\raggedright\arraybackslash}p{3.4cm}>{\centering\arraybackslash}p{1.6cm}>{\centering\arraybackslash}p{1.6cm}>{\centering\arraybackslash}p{1.6cm}>{\raggedright\arraybackslash}X}
\toprule
\textbf{Modality group} & \textbf{Representative source datasets} & \textbf{Anatomy} & \textbf{Abnormal} & \textbf{Disease} & \textbf{Clinical scope} \\
\midrule
CXR & MIMIC-CXR, IU-Xray, CheXpert Plus, Medical-Diff-VQA, MMXU & 14 & 30 & 9 & Thoracic structures, pleural findings, pulmonary parenchymal abnormalities, mediastinal and cardiovascular findings. \\
CT & CT-RATE, AMOS-MM, BIMCV-R & 19 & 22 & 10 & Chest, abdominal, pelvic, vascular, hepatic, renal, and mediastinal findings across volumetric CT studies. \\
BRAIN & SSPH & 2 & 9 & 7 & Brain parenchymal and ventricular/cisternal abnormalities, including infarction, hemorrhage, soft-tissue lesions, infection, demyelination, and trauma-related entities. \\
US & CURG & 7 & 8 & 2 & Breast, thyroid, liver, gallbladder, axillary, vascular, and ductal findings in organ-specific ultrasound studies. \\
\bottomrule
\end{tabularx}
\end{table*}

We organize clinical entities into three granularities---anatomical structures, abnormal findings, and pathological conditions---and group them by four modality families: CXR, CT, BRAIN, and US. The modality-level coverage is summarized in Table~\ref{tab:entity_group_summary}; the complete entity vocabulary is listed below.

\subsection{CXR entities}
\begin{description}
\item[Anatomy (14)] Pulmonary Parenchyma; Bronchi (Lower Respiratory Tract); Pulmonary Interstitium; Trachea\_Tracheal Lumen; Cardiac Chambers; Cardiac Valves; Pulmonary Vasculature; Mediastinum; Pleura; Diaphragm; Stomach; Ribs; Sternum; Aorta.
\item[Abnormal findings (30)] Diffusely distributed, multiple pulmonary nodules and masses (sarcoidosis, metastasis, pneumoconiosis) in Pulmonary Parenchyma; Bone deformity in Ribs; Interstitial fibrosis and thickening (e.g., reticular changes) in Pulmonary Interstitium; Cardiac wall hypertrophy (e.g., myocardial hypertrophy) in Cardiac Walls; Ground-glass opacity (e.g., slightly high-density shadow due to inflammation) in Pulmonary Parenchyma; Bone fracture in Ribs; Lung consolidation (e.g., high-density shadow due to inflammation) in Pulmonary Parenchyma; Pulmonary fibrosis in Pulmonary Parenchyma; Venous dilatation in Pulmonary Vasculature; Callus and post-fracture healing changes in Ribs; Bone deformity in Thoracic spine; Bone fracture in Thoracic spine; Soft tissue density lesions in the mediastinum (including mediastinal tumors) in Mediastinum; Arterial dilatation in Aorta; Mediastinal shift in Mediastinum; Solitary pulmonary nodule and mass (peripheral lung cancer, granuloma) in Pulmonary Parenchyma; Arterial dilatation in Pulmonary Vasculature; Cardiac chamber enlargement (atrial or ventricular enlargement) in Cardiac Chambers; Atelectasis in Pulmonary Parenchyma; Atherosclerotic plaque of the arterial wall in Aorta; Abnormal arterial course in Aorta; Bronchiectasis in Bronchi (Lower Respiratory Tract); Lymph node enlargement in the mediastinum in Mediastinum; Pericardial effusion in Pericardium; Bone fracture in Sternum; Eventration of the diaphragm in Diaphragm; Hernia or abnormal course of the gastrointestinal tract in Stomach; Pleural thickening in Pleura; Pleural effusion (including localized effusion) in Pleura; Venous dilatation in Thoracic veins.
\item[Diseases (9)] Aortic aneurysm; Aortosclerosis; Aortic valve disease; Cardiomyopathy; Pulmonary hypertension; Pneumonia; Pulmonary tuberculosis; Pneumomediastinum; Skeletal deformities.
\end{description}

\subsection{CT entities}
\begin{description}
\item[Anatomy (19)] Pleura; Mediastinum; Cardiac Chambers; Pericardium; Coronary Arteries; Pulmonary Parenchyma; Bronchi (Lower Respiratory Tract); Pulmonary Interstitium; Breast Tissue; Chest Wall Soft Tissues; Thoracic Vertebrae; Thoracic Arteries; Liver\_Hepatic Parenchyma; Esophagus; Adrenal Glands; Gallbladder Lumen; Gallbladder Morphology; Spleen\_Splenic Parenchyma; Kidneys\_Renal Parenchyma.
\item[Abnormal findings (22)] Lymphadenopathy within the mediastinum in Mediastinum; Emphysema in Pulmonary Parenchyma; Solitary pulmonary nodule and mass (peripheral lung cancer, granuloma) in Pulmonary Parenchyma; Diffusely distributed, multiple pulmonary nodules and masses (sarcoidosis, metastasis, pneumoconiosis) in Pulmonary Parenchyma; Abnormal esophageal position in Esophagus; Pulmonary fibrosis in Pulmonary Parenchyma; Soft tissue density lesion of the pleura in Pleura; Arterial dilation in Thoracic Arteries; Arterial wall atherosclerotic plaque in Thoracic Arteries; Ground-glass opacity (e.g., slightly high-density shadow due to inflammation) in Pulmonary Parenchyma; Pleural thickening in Pleura; Pleural effusion (including loculated effusion) in Pleura; Renal calcification in Kidneys\_Renal Parenchyma; Lung consolidation (e.g., high-density shadow due to inflammation) in Pulmonary Parenchyma; Pericardial effusion in Pericardium; Soft tissue density lesion within the mediastinum (including mediastinal tumors) in Mediastinum; Bone deformity in Thoracic Vertebrae; Decreased hepatic parenchymal density (e.g., fatty change) in Liver\_Hepatic Parenchyma; Interstitial fibrosis and thickening (e.g., reticular changes) in Pulmonary Interstitium; Cystic lesion of the renal parenchyma in Kidneys\_Renal Parenchyma; Atelectasis in Pulmonary Parenchyma; Abnormal hepatic parenchymal morphology (including contour, size, hepatic lobe ratio, width of hepatic fissures, etc.) in Liver\_Hepatic Parenchyma.
\item[Diseases (10)] Pneumonia (e.g., viral pneumonia, bacterial pneumonia, mycoplasma pneumonia, etc.); Chronic obstructive pulmonary disease\_Emphysema; Pulmonary hypertension; Bronchitis; Interstitial lung disease; Coronary heart disease\_Coronary atherosclerotic heart disease; Aortic aneurysm; Cholelithiasis; Hiatal hernia; Thoracic spondylosis.
\end{description}

\subsection{BRAIN entities}
\begin{description}
\item[Anatomy (2)] Brain Parenchyma (including Cerebrum and Cerebellum); Ventricles and Cisterns.
\item[Abnormal findings (9)] Intraventricular and intracisternal hemorrhage in Ventricles and Cisterns; Other non-mass brain parenchymal lesions (such as mineral deposition due to metabolic encephalopathy) in Brain Parenchyma (including Cerebrum and Cerebellum); Brain parenchymal hyperdense foci in Brain Parenchyma (including Cerebrum and Cerebellum); Cerebral softening lesions in Brain Parenchyma (including Cerebrum and Cerebellum); Cerebral infarction in Brain Parenchyma (including Cerebrum and Cerebellum); Cerebral ischemic lesions in Brain Parenchyma (including Cerebrum and Cerebellum); Intraventricular and intracisternal soft tissue density lesions in Ventricles and Cisterns; Brain parenchymal soft tissue density lesions in Brain Parenchyma (including Cerebrum and Cerebellum); Brain parenchymal hemorrhage and contusion in Brain Parenchyma (including Cerebrum and Cerebellum).
\item[Diseases (7)] Cerebrovascular malformations (e.g., cavernous hemangioma, arteriovenous malformation); Intracranial infections (e.g., encephalitis, brain abscess); Traumatic lesions (e.g., cerebral contusion, diffuse axonal injury); Metastatic tumors (metastases); Demyelinating diseases (e.g., multiple sclerosis); Intracranial hemorrhage (e.g., subdural hematoma, intracerebral hematoma); Cerebral small vessel disease (white matter demyelination).
\end{description}

\subsection{US entities}
\begin{description}
\item[Anatomy (7)] Gallbladder; Breast tissue; Thyroid; Hepatic parenchyma; Axilla; Veins; Mammary ducts.
\item[Abnormal findings (8)] Mammary duct ectasia in Mammary ducts; Increased hepatic parenchymal density (such as calcification, iron deposition) in Hepatic parenchyma; Decreased hepatic parenchymal density (such as fatty degeneration) in Hepatic parenchyma; Lymphadenopathy in Axilla; Breast soft tissue density lesions in Breast tissue; Breast cystic lesions in Breast tissue; Hepatic parenchymal morphological abnormalities (including contour, size, hepatic lobe ratio, hepatic fissure width, etc.) in Hepatic parenchyma; Thyroid soft tissue density lesions in Thyroid.
\item[Diseases (2)] Breast cancer; Mammary duct ectasia.
\end{description}

\section{Supplementary Prompt Templates}
\label{sec:supp_prompts}

This section provides the prompt templates used for report decomposition, comparative VQA generation, conversion to open-ended QA, and image-dependency filtering. Placeholders enclosed in braces or square brackets are replaced with the corresponding entity name, anatomy name, abnormality name, disease term, report text, or question during data construction.

\subsection{Report-decomposition prompts}
\label{sec:report_decomposition_prompts}
\begin{promptbox}{Prompt S1. Anatomy-specific report decomposition}
You are a board-certified radiologist. Your task is to extract descriptive information pertaining to a specific anatomical structure ({anatomy}) from the provided medical imaging diagnostic report. Please adhere to the following guidelines:

- Extract with Precision: Identify and extract only the information that directly references the {anatomy}, including any noted abnormalities, lesions, or pathological findings associated with it.

- Specify Anatomical Details: If the report mentions particular regions, subparts, or specific anatomical features of the {anatomy}, ensure these are included in your description. This may involve affected areas, normal characteristics, or other notable observations.

- Be Concise and Direct: Present the extracted information in a clear and straightforward manner, using original wording from the report. Avoid adding interpretations, summaries, or external context.

- Formatting Requirements: Return only a single block of text containing the consolidated descriptive information related to the {anatomy}. Do not include prefixes, headings, bullet points, or any additional commentary. Even if multiple distinct sections or features are described, integrate them into one coherent paragraph using standard medical terminology and reporting conventions.
\end{promptbox}

\begin{promptbox}{Prompt S2. Disease-specific report decomposition}
You are a board-certified radiologist. Your task is to extract descriptive information related to the specific condition "{disease term}" from the provided medical imaging diagnostic report. Please adhere to the following guidelines:

- Precise Extraction: Identify and extract only information that directly references "{disease term}", including any documented findings associated with it.

- Verbatim Retention: Use the original wording from the report. Present the extracted information clearly and directly, without adding interpretation, summary, or external context.

- Integrated Output: Return only a single, unified text block containing all extracted descriptive information related to "{disease term}". Use standard medical terminology and reporting conventions to combine distinct observations or features into one coherent paragraph.

- Strict Formatting: Do not include prefixes, headings, bullet points, or any additional commentary.
\end{promptbox}

\begin{promptbox}{Prompt S3. Abnormality-specific report decomposition}
You are a professional radiologist. I will provide you with the following information:

- The name of a specific anatomical structure: {anatomy_name}
- A descriptive text from a real medical report regarding this structure
- The name of a potential abnormality associated with this structure: {abnormality}

Please complete the following task:

Strictly extract from the provided medical report all content related to the abnormality {abnormality}. Adhere to the following principles during extraction:

- Focus only on content directly pertaining to the presence, morphology, location, size, or characteristics of {abnormality};
- You may paraphrase the original text appropriately to ensure coherence and completeness, but do not add any information not mentioned in the report;
- If the abnormality is not mentioned at all in the report, the output value should be null;
- The final output should consist of 1--4 complete sentences forming a coherent "overall description of the abnormality." Do not use fragmented phrases or isolated keywords.

Provide the output in the following JSON format:

{
  "abnormality description": "xxx"  // or null
}

Important requirements:

- Strictly base the extraction on the original report text. Do not make any inferences, assumptions, or incorporate external knowledge;
- If the report explicitly states the absence of the abnormality (e.g., "no evidence of ..."), extract such statements truthfully without omission;
- Maintain medical accuracy and professionalism in the description.
\end{promptbox}

\subsection{Comparative VQA prompts}
\label{sec:comparative_vqa_prompts}
\begin{promptbox}{Prompt S4. Comparative multiple-choice VQA generation}
As a professional radiologist, you will receive two imaging reports from different cases. Both reports describe a specific abnormality within the anatomical region of [anatomy]. To assist your comprehensive understanding of the imaging findings, I will also provide the overall imaging descriptions of this anatomical region from both reports for contextual reference. Your task is to design several multiple-choice questions based on the specific imaging characteristics of the [abnormal] finding, focusing on comparing its manifestations between the two images.

Specific requirements are as follows:

1. Question Focus: Questions should focus specifically on the differential manifestations of the abnormality [abnormal] between the two images. Aspects for comparison include, but are not limited to: presence/absence, location, size, morphology, margins, internal characteristics, associated findings, or potential etiology. The overall imaging appearance of the [anatomy] may be used as auxiliary reference context where appropriate.
2. Comprehensive Coverage: Questions should cover all identifiable dimensions of difference, ensuring variety and that no key comparative points are omitted.
3. Answer and Rationale: Each question must include the correct answer (e.g., A, B, C, D) accompanied by a concise rationale explaining the basis for the judgment.
4. Perspective of Expression: The wording of questions and rationales should adopt an "image comparison" perspective. References to the cases must be specific; for example, use "the imaging from Case A" and "the imaging from Case B," avoiding generic terms like "one image" or "the other image." Additionally, refrain from using phrases such as "the report indicates," reflecting a scenario where judgment is based solely on the imaging information itself.
5. Relevance and Specificity: All questions must strictly pertain to the actual differences present between the two images. Avoid content related to unchanged features or aspects irrelevant to the comparison.
6. Information Fidelity: Questions, options, answers, and rationales must be strictly based on the provided factual information. Do not fabricate or introduce details not mentioned.

Output Format (JSON):
Each question must strictly adhere to the following JSON structure:
{
  "question": "Question content",
  "condition": "[anatomy]_and_[abnormal]",
  "content_type": "Type of difference, e.g., Presence, Location, Size, etc.",
  "options": {
    "A": "Option A content",
    "B": "Option B content",
    "C": "Option C content",
    "D": "Option D content"
  },
  "answer": "Correct answer (e.g., A)",
  "reason": "Brief explanation for the answer"
}
\end{promptbox}

\begin{promptbox}{Prompt S5. Conversion from multiple-choice to open-ended QA}
You will receive a multiple-choice question about two medical images, which includes the question stem, answer options, and an explanation. Your task is to convert this question into a short-answer format (question-answer pair) while strictly adhering to the image comparison perspective.

Rules:

- Expression Guidelines:
  - Use an "image comparison" perspective in both the question and the answer.
  - Explicitly reference the case source, e.g., "The image from Case A shows...", "The image from Case B demonstrates...". Avoid vague terms like "one image" or "a certain scan".
  - Avoid phrases such as "the report indicates" to ensure interpretation is based solely on the image information.

- Content Control:
  - Do not introduce external information or subjective inferences. Base the response exclusively on the provided image details.

Output Format:
The response must be in the following JSON format:
[
  {
    "question": "Your generated question here",
    "answer": "The correct short answer here"
  }
]
\end{promptbox}

\begin{promptbox}{Prompt S6. Image-dependency filtering}
You are a professional radiologist. You will be given a question below. Please determine whether answering this question requires examining two medical images. If two images are strictly necessary to answer the question, respond with "yes"; otherwise, respond with "no".

Instructions:
- Carefully analyze the question to identify whether it involves comparison, change over time, bilateral assessment, or any other reasoning that inherently requires two distinct images.
- Questions that can be answered from a single image (e.g., identifying a finding, describing anatomy, or assessing a single region) should return "no".
- Questions involving comparison between two studies (e.g., pre- vs. post-treatment, left vs. right side shown in separate images, follow-up vs. baseline) should return "yes".
- Respond with only "yes" or "no" -- no explanation or additional text.

Question: {question}
Answer:
\end{promptbox}

\section{Supplementary Model and Training Details}
\label{sec:supp_training}

\begin{table}[!htbp]
\centering
\small
\caption{Training hyperparameters for the two-stage MedReCo-VLM training procedure.}
\label{tab:training_hyperparameters}
\suppTableTopSpace
\begin{tabularx}{\textwidth}{>{\raggedright\arraybackslash}p{3.0cm}>{\raggedright\arraybackslash}X>{\raggedright\arraybackslash}X}
\toprule
\textbf{Hyperparameter} & \textbf{Stage 1: pre-training} & \textbf{Stage 2: instruction tuning} \\
\midrule
Optimizer & AdamW & AdamW \\
Learning rate & $1\times10^{-4}$ & $2\times10^{-5}$ (LLM) / $1\times10^{-4}$ (projection) \\
Learning-rate schedule & Linear warm-up + cosine decay & Linear warm-up + cosine decay \\
Batch size & 14 (3D) / 50 (2D) per GPU & 64 (global) \\
Epochs & 10 & 1 \\
Input resolution & $240\!\times\!480\!\times\!480$ (3D) / $480\!\times\!480$ (2D) & -- \\
Maximum sequence length & -- & 2048 \\
Frozen modules & -- & $\phi_{\mathrm{vis}}$, $\phi_{\mathrm{text}}$ \\
Hardware & $8\times$ A100 (80 GB) & $8\times$ A100 (80 GB) \\
Training time & $\sim$72 h & $\sim$72 h \\
\bottomrule
\end{tabularx}
\end{table}

\section{Supplementary Evaluation Protocols and Results}
\label{sec:supp_results_tables}

\subsection{Evaluation protocols}
\label{sec:supp_eval}

For controllable image retrieval, each query consists of a query image and a target entity condition. The model ranks candidate images from the same modality group according to entity-specific similarity. For generative comparative interpretation, each query consists of two medical images and a target entity condition; the model generates a natural-language comparison describing entity-specific similarities, differences, or interval changes. Table~\ref{tab:evaluation_protocols} summarizes the evaluation settings.

\begin{table*}[!htbp]
\centering
\small
\caption{Evaluation protocols and primary metrics.}
\label{tab:evaluation_protocols}
\suppTableTopSpace
\begin{tabularx}{\textwidth}{>{\raggedright\arraybackslash}p{3.0cm}>{\raggedright\arraybackslash}X>{\raggedright\arraybackslash}X>{\raggedright\arraybackslash}p{3.2cm}}
\toprule
\textbf{Evaluation setting} & \textbf{Query / input} & \textbf{Candidate / reference} & \textbf{Metrics} \\
\midrule
Internal retrieval & Query image and entity from an internal validation split. & Candidate gallery from the same institution and modality group. & Recall@1, Recall@3, Recall@5. \\
External retrieval & Query image and entity from a held-out external dataset. & Candidate gallery from the same held-out dataset. & Recall@1, Recall@3, Recall@5. \\
Cross-centre retrieval & Query image and entity from an internal test split. & Candidate gallery from a non-overlapping held-out external centre. & Recall@1, Recall@3, Recall@5. \\
Confusable differential retrieval & Query image and clinically confusable target entity. & Candidate gallery restricted to visually or anatomically confusable entities. & Recall@1 and macro-average gain over the strongest applicable baseline. \\
Broad paired-image VQA & Cross-patient image pair and target entity. & Closed-ended answer or open-ended reference answer. & Accuracy; BLEU; METEOR; BERTScore; RaTEScore; RadGraph F1. \\
Longitudinal follow-up VQA & Same-patient prior--current pair and target finding/attribute. & Change label and reference comparative answer. & Accuracy and attribute-stratified accuracy. \\
\bottomrule
\end{tabularx}
\end{table*}

\subsection{Complete numerical results}
All values are percentages unless otherwise specified. For tables with confidence intervals, values are reported as mean [95\% confidence interval]. $\Delta$ denotes the absolute percentage-point difference between \ours{} and the best applicable baseline.

\begin{table*}[!htbp]
\centering
\scriptsize
\setlength{\tabcolsep}{2.5pt}
\renewcommand{\arraystretch}{1.10}
\caption{Internal validation of entity-conditioned reference-case retrieval (Figure~2a).}
\label{tab:supp_retrieval_internal_compact}
\suppTableTopSpace
\resizebox{\textwidth}{!}{%
\begin{tabular}{llllllllllll}
\toprule
Modality & Entity & CT-CLIP & MedCLIP & PMC-CLIP & BiomedCLIP & HLIP & \ourretrieval{} & Best baseline & $\Delta$ & $p$ & Sig. \\
\midrule
Chest X-ray & Abnormality & -- & 40.61 [39.19--42.00] & 45.41 [43.98--46.84] & 40.17 [38.76--41.60] & -- & 48.44 [47.01--49.87] & PMC-CLIP & +3.03 & $3.11\times 10^{-5}$ & *** \\
Chest X-ray & Anatomy & -- & 63.35 [62.19--64.51] & 63.31 [62.14--64.44] & 64.07 [62.91--65.24] & -- & 70.21 [69.10--71.31] & BiomedCLIP & +6.14 & $7.27\times 10^{-33}$ & *** \\
Chest X-ray & Disease & -- & 64.00 [62.22--65.68] & 65.68 [63.97--67.36] & 64.10 [62.39--65.78] & -- & 70.85 [69.20--72.46] & PMC-CLIP & +5.17 & $5.68\times 10^{-12}$ & *** \\
CT & Abnormality & 47.88 [46.59--49.19] & -- & -- & -- & -- & 51.38 [50.07--52.67] & CT-CLIP & +3.50 & $6.75\times 10^{-8}$ & *** \\
CT & Anatomy & 65.01 [63.60--66.38] & -- & -- & -- & -- & 69.39 [68.00--70.74] & CT-CLIP & +4.38 & $1.61\times 10^{-11}$ & *** \\
CT & Disease & 60.98 [59.42--62.52] & -- & -- & -- & -- & 65.67 [64.13--67.18] & CT-CLIP & +4.69 & $8.83\times 10^{-12}$ & *** \\
Brain MRI & Abnormality & -- & -- & -- & -- & 30.78 [29.36--32.23] & 36.09 [34.60--37.56] & HLIP & +5.31 & $8.03\times 10^{-9}$ & *** \\
Brain MRI & Anatomy & -- & -- & -- & -- & 53.13 [49.89--56.36] & 63.62 [60.49--66.74] & HLIP & +10.49 & $1.13\times 10^{-10}$ & *** \\
Brain MRI & Disease & -- & -- & -- & -- & 70.93 [69.42--72.44] & 77.07 [75.67--78.49] & HLIP & +6.14 & $1.07\times 10^{-17}$ & *** \\
US & Abnormality & -- & -- & 48.89 [45.98--51.80] & 50.51 [47.60--53.42] & -- & 52.65 [49.74--55.57] & BiomedCLIP & +2.14 & 0.0889 & n.s. \\
US & Anatomy & -- & -- & 68.31 [65.48--71.13] & 68.12 [65.39--71.04] & -- & 70.58 [67.85--73.41] & PMC-CLIP & +2.27 & 0.0278 & * \\
US & Disease & -- & -- & 70.78 [63.64--77.92] & 72.73 [65.58--79.87] & -- & 81.17 [74.68--87.01] & BiomedCLIP & +8.44 & 0.0106 & * \\
\bottomrule
\end{tabular}%
}
\vspace{2pt}
\suppTableBottomSpace
\begin{minipage}{\textwidth}\footnotesize\emph{Note.} Values are Recall@1. Significance markers compare \ourretrieval{} with the strongest applicable baseline: $^{*}p<0.05$, $^{**}p<0.01$, $^{***}p<0.001$; n.s., not significant.\end{minipage}
\end{table*}

\begin{table*}[!htbp]
\centering
\scriptsize
\setlength{\tabcolsep}{3.5pt}
\renewcommand{\arraystretch}{1.10}
\caption{Fine-grained retrieval analysis on (i) CT disease entities and (ii) Brain MRI abnormal findings, corresponding to Figure~2b.}
\label{tab:supp_retrieval_finegrained_compact}
\suppTableTopSpace
\resizebox{\textwidth}{!}{%
\begin{tabular}{llccccll}
\toprule
Modality & Category & CT-CLIP & HLIP & \ourretrieval{} & Best baseline & $\Delta$ \\
\midrule
CT & Pneumonia & 84.88 & -- & 90.41 & CT-CLIP & $+$5.53 \\
CT & Interstitial Lung Disease & 85.90 & -- & 87.96 & CT-CLIP & $+$2.06 \\
CT & Bronchitis & 73.97 & -- & 84.25 & CT-CLIP & $+$10.28 \\
CT & Pulmonary Hypertension & 59.38 & -- & 74.38 & CT-CLIP & $+$15.00 \\
CT & Cholelithiasis & 53.26 & -- & 67.39 & CT-CLIP & $+$14.13 \\
CT & Thoracic Spondylosis & 55.51 & -- & 59.03 & CT-CLIP & $+$3.52 \\
CT & COPD / Emphysema & 50.82 & -- & 54.45 & CT-CLIP & $+$3.63 \\
CT & Aortic Aneurysm & 48.28 & -- & 50.19 & CT-CLIP & $+$1.91 \\
CT & Coronary Heart Disease & 44.20 & -- & 50.13 & CT-CLIP & $+$5.93 \\
CT & Hiatal Hernia & 39.86 & -- & 40.82 & CT-CLIP & $+$0.96 \\
CT & Average & 59.61 & -- & 65.90 & CT-CLIP & $+$6.29 \\
\midrule
Brain MRI & Intraventr.\,/\,intracist. soft tissue lesions & -- & 32.98 & 51.06 & HLIP & $+$18.08 \\
Brain MRI & Parenchymal hyperdense foci & -- & 31.77 & 40.39 & HLIP & $+$8.62 \\
Brain MRI & Intraventr.\,/\,intracist. hemorrhage & -- & 33.93 & 39.29 & HLIP & $+$5.36 \\
Brain MRI & Parenchymal hemorrhage \& contusion & -- & 33.54 & 37.89 & HLIP & $+$4.35 \\
Brain MRI & Parenchymal soft tissue density lesions & -- & 31.45 & 35.75 & HLIP & $+$4.30 \\
Brain MRI & Other non-mass parenchymal lesions & -- & 30.13 & 35.15 & HLIP & $+$5.02 \\
Brain MRI & Cerebral softening lesions & -- & 28.67 & 34.97 & HLIP & $+$6.30 \\
Brain MRI & Cerebral ischemic lesions & -- & 29.69 & 33.98 & HLIP & $+$4.29 \\
Brain MRI & Cerebral infarction & -- & 31.45 & 33.06 & HLIP & $+$1.61 \\
Brain MRI & Average & -- & 31.51 & 37.95 & HLIP & $+$6.44 \\
\bottomrule
\end{tabular}%
}
\suppTableBottomSpace
\begin{minipage}{\textwidth}\footnotesize\emph{Note.} Values are Recall@1 (\%). Category-level point estimates are reported without confidence intervals or $p$ values. For each modality only the modality-specific baseline is shown (CT-CLIP for CT, HLIP for Brain MRI); ``--'' indicates not applicable. Rows are sorted by \ourretrieval{} performance (descending) within each modality; the bottom row of each block is the macro-average over its entities. $\Delta$ = \ours{} $-$ best baseline.\end{minipage}
\end{table*}

\begin{table*}[!htbp]
\centering
\scriptsize
\setlength{\tabcolsep}{2.5pt}
\renewcommand{\arraystretch}{1.10}
\caption{External and cross-dataset entity-conditioned retrieval performance (Figure~2c--d).}
\label{tab:supp_retrieval_external_crossdataset_compact}
\suppTableTopSpace
\resizebox{\textwidth}{!}{%
\begin{tabular}{llllllllllll}
\toprule
Setting & Modality & Entity & CT-CLIP & MedCLIP & PMC-CLIP & BiomedCLIP & \ourretrieval{} & Best baseline & $\Delta$ & $p$ & Sig. \\
\midrule
External & Chest X-ray & Anatomy & -- & 65.17 [63.86--66.50] & 64.37 [63.08--65.70] & 62.02 [60.70--63.35] & 71.54 [70.31--72.79] & PMC-CLIP & +7.17 & $7.25\times 10^{-36}$ & *** \\
External & Chest X-ray & Abnormality & -- & 42.47 [40.94--44.03] & 34.24 [32.76--35.75] & 39.59 [38.03--41.12] & 50.41 [48.82--52.02] & MedCLIP & +7.94 & $1.32\times 10^{-22}$ & *** \\
External & Chest X-ray & Disease & -- & 64.47 [62.70--66.20] & 58.33 [56.57--60.10] & 62.77 [61.00--64.50] & 68.57 [66.90--70.20] & MedCLIP & +4.10 & $1.26\times 10^{-7}$ & *** \\
External & CT & Anatomy & 77.13 [74.94--79.20] & -- & -- & -- & 83.91 [82.11--85.66] & CT-CLIP & +6.78 & $8.31\times 10^{-10}$ & *** \\
External & CT & Abnormality & 34.34 [32.21--36.42] & -- & -- & -- & 40.52 [38.34--42.70] & CT-CLIP & +6.18 & $2.36\times 10^{-6}$ & *** \\
External & CT & Disease & 76.34 [73.92--78.77] & -- & -- & -- & 80.16 [77.82--82.41] & CT-CLIP & +3.82 & $1.81\times 10^{-3}$ & ** \\
Cross-dataset & Chest X-ray & Anatomy & -- & 41.31 [40.09--42.52] & 40.27 [39.02--41.47] & 42.64 [41.37--43.89] & 55.38 [54.13--56.63] & BiomedCLIP & +12.74 & $3.44\times 10^{-58}$ & *** \\
Cross-dataset & Chest X-ray & Abnormality & -- & 47.61 [45.72--49.53] & 47.12 [45.27--49.04] & 48.70 [46.81--50.62] & 55.60 [53.71--57.48] & BiomedCLIP & +6.90 & $7.29\times 10^{-10}$ & *** \\
Cross-dataset & Chest X-ray & Disease & -- & 39.95 [38.18--41.75] & 43.06 [41.22--44.86] & 41.22 [39.39--43.06] & 57.61 [55.78--59.41] & PMC-CLIP & +14.55 & $1.06\times 10^{-35}$ & *** \\
Cross-dataset & CT & Anatomy & 49.57 [47.86--51.28] & -- & -- & -- & 62.13 [60.45--63.78] & CT-CLIP & +12.56 & $4.84\times 10^{-26}$ & *** \\
Cross-dataset & CT & Abnormality & 34.28 [31.54--37.01] & -- & -- & -- & 38.60 [35.78--41.52] & CT-CLIP & +4.32 & 0.0115 & * \\
Cross-dataset & CT & Disease & 58.31 [55.64--60.98] & -- & -- & -- & 64.71 [62.12--67.23] & CT-CLIP & +6.40 & $8.80\times 10^{-5}$ & *** \\
\bottomrule
\end{tabular}%
}
\suppTableBottomSpace
\begin{minipage}{\textwidth}\footnotesize\emph{Note.} External validation evaluates queries and candidates from the same held-out institution; cross-dataset retrieval uses queries and galleries from non-overlapping institutions.\end{minipage}
\end{table*}

\begin{table*}[!htbp]
\centering
\scriptsize
\setlength{\tabcolsep}{2.5pt}
\renewcommand{\arraystretch}{1.10}
\caption{Clinically confusable differential retrieval groups (Figure~3a).}
\label{tab:supp_differentials_compact}
\suppTableTopSpace
\resizebox{\textwidth}{!}{%
\begin{tabular}{cllcclcccccc}
\toprule
ID & Differential group & Modality & Classes & $n$ & Best baseline & Baseline (95\% CI) & \ourretrieval{} (95\% CI) & $\Delta$ & $b/c$ & mid-$p$ \\
\midrule
1 & Mediastinal lymphadenopathy vs Vascular structure/aneurysm vs Mediastinal cyst & CT & 3 & 75 & CT-CLIP & 77.33 (66.7--85.3) & 86.67 (77.2--92.6) & +9.34 & 12/5 & $0.0481^{*}$ \\
2 & Hydronephrosis vs Parapelvic/renal sinus cyst & CT & 2 & 29 & CT-CLIP & 65.52 (47.3--80.1) & 82.76 (65.4--92.4) & +17.24 & 6/1 & $0.0352^{*}$ \\
3 & Simple hepatic cyst vs Hepatic hemangioma vs Hepatic metastasis & CT & 3 & 119 & CT-CLIP & 47.90 (39.1--56.8) & 58.82 (49.8--67.3) & +10.92 & 30/17 & $0.0297^{*}$ \\
4 & Atelectasis vs Pneumonia consolidation vs Obstructing central tumor & CT & 3 & 995 & CT-CLIP & 52.26 (49.1--55.3) & 57.29 (54.2--60.3) & +5.03 & 238/188 & $0.0077^{**}$ \\
5 & Emphysema vs Pulmonary bulla vs Pulmonary cyst & CT & 3 & 351 & CT-CLIP & 77.78 (73.1--81.8) & 79.49 (75.0--83.4) & +1.71 & 31/25 & $0.2135$ \\
6 & Lung cancer vs Granuloma vs Nipple shadow & Chest X-ray & 3 & 124 & PMC-CLIP & 79.03 (71.1--85.3) & 86.29 (79.1--91.3) & +7.26 & 13/4 & $0.0154^{*}$ \\
\bottomrule
\end{tabular}%
}
\suppTableBottomSpace
\begin{minipage}{\textwidth}\footnotesize\emph{Note.} Recall@1 reported as percentages with 95\% confidence intervals; $n$ is the number of retrieval queries. $b/c$ are the discordant pairs (\ourretrieval{} correct\,/\,baseline correct) from McNemar's exact test, and mid-$p$ is the corresponding two-sided value. ${}^{*}p<0.05$, ${}^{**}p<0.01$. ``Baseline'' is the best-performing baseline model for each group.\end{minipage}
\end{table*}

\begin{table*}[!htbp]
\centering
\scriptsize
\setlength{\tabcolsep}{2.5pt}
\renewcommand{\arraystretch}{1.10}
\caption{In-domain generative comparative interpretation performance across broad image pairs (Figure~4a).}
\label{tab:supp_comparative_indomain_compact}
\suppTableTopSpace
\resizebox{\textwidth}{!}{%
\begin{tabular}{llllllllllllll}
\toprule
Modality & Dataset & Entity & Metric & RadFM & CheXagent & MedGemma & Hulu-Med & Lingshu & \ours{} & Best baseline & $\Delta$ & $p$ & Sig. \\
\midrule
Chest X-ray & MIMIC-CXR & Abnormality & Accuracy & 25.25 [23.70--26.86] & 24.44 [23.54--25.36] & 42.54 [41.49--43.59] & 51.19 [50.25--52.13] & 57.28 [56.23--58.32] & 70.95 [70.09--71.80] & Lingshu-7B & +13.67 & $3.30\times 10^{-113}$ & *** \\
Chest X-ray & MIMIC-CXR & Abnormality & RaTEScore & 27.50 [26.93--28.09] & 39.71 [39.23--40.19] & 54.70 [54.37--55.04] & 56.96 [56.59--57.34] & 55.42 [55.05--55.81] & 64.05 [63.62--64.48] & Hulu-Med-32B & +7.09 & 0.0001 & *** \\
Chest X-ray & MIMIC-CXR & Anatomy & Accuracy & 30.88 [25.80--36.47] & 25.03 [22.43--27.82] & 52.07 [48.96--55.17] & 57.11 [54.01--60.16] & 65.05 [62.03--67.96] & 72.45 [69.59--75.14] & Lingshu-7B & +7.40 & $8.32\times 10^{-6}$ & *** \\
Chest X-ray & MIMIC-CXR & Anatomy & RaTEScore & 25.27 [23.62--26.86] & 37.10 [35.56--38.64] & 53.67 [52.67--54.63] & 56.61 [55.34--57.90] & 54.81 [53.67--55.98] & 62.49 [61.04--63.96] & Hulu-Med-32B & +5.88 & 0.0001 & *** \\
Chest X-ray & MIMIC-CXR & Disease & Accuracy & 30.77 [28.52--33.11] & 32.72 [31.24--34.24] & 37.38 [35.85--38.94] & 50.40 [48.80--52.00] & 54.53 [52.94--56.12] & 62.00 [60.44--63.54] & Lingshu-7B & +7.47 & $1.54\times 10^{-16}$ & *** \\
Chest X-ray & MIMIC-CXR & Disease & RaTEScore & 14.37 [13.67--15.08] & 27.57 [26.89--28.29] & 50.74 [50.16--51.32] & 54.12 [53.36--54.85] & 50.96 [50.30--51.58] & 57.64 [56.81--58.47] & Hulu-Med-32B & +3.51 & 0.0001 & *** \\
CT & CT-RATE & Abnormality & Accuracy & 23.40 [21.36--25.58] & -- & -- & 45.40 [43.27--47.55] & -- & 69.32 [68.17--70.46] & Hulu-Med-32B & +23.92 & $4.52\times 10^{-59}$ & *** \\
CT & CT-RATE & Abnormality & RaTEScore & 28.01 [27.30--28.73] & -- & -- & 53.15 [52.14--54.14] & -- & 61.42 [60.41--62.40] & Hulu-Med-32B & +8.26 & 0.0001 & *** \\
CT & CT-RATE & Anatomy & Accuracy & 36.15 [32.86--39.59] & -- & -- & 36.86 [34.84--38.94] & -- & 62.34 [60.25--64.37] & Hulu-Med-32B & +25.47 & $1.92\times 10^{-65}$ & *** \\
CT & CT-RATE & Anatomy & RaTEScore & 18.54 [17.53--19.57] & -- & -- & 56.90 [55.71--58.06] & -- & 66.47 [65.46--67.49] & Hulu-Med-32B & +9.58 & 0.0001 & *** \\
CT & CT-RATE & Disease & Accuracy & 29.93 [27.53--32.45] & -- & -- & 36.14 [33.82--38.53] & -- & 53.05 [50.59--55.49] & Hulu-Med-32B & +16.91 & $2.30\times 10^{-22}$ & *** \\
CT & CT-RATE & Disease & RaTEScore & 13.87 [13.12--14.62] & -- & -- & 54.03 [52.68--55.35] & -- & 60.29 [59.03--61.49] & Hulu-Med-32B & +6.26 & 0.0001 & *** \\
Brain MRI & Brain MRI & Abnormality & Accuracy & 23.24 [21.10--25.53] & -- & -- & 42.74 [41.20--44.30] & -- & 63.58 [62.06--65.08] & Hulu-Med-32B & +20.84 & $1.62\times 10^{-81}$ & *** \\
Brain MRI & Brain MRI & Abnormality & RaTEScore & 21.15 [20.39--21.91] & -- & -- & 48.82 [48.03--49.62] & -- & 65.88 [65.22--66.55] & Hulu-Med-32B & +17.07 & 0.0001 & *** \\
Brain MRI & Brain MRI & Anatomy & Accuracy & 30.72 [26.62--35.15] & -- & -- & 45.64 [42.89--48.41] & -- & 68.29 [65.66--70.82] & Hulu-Med-32B & +22.66 & $7.87\times 10^{-34}$ & *** \\
Brain MRI & Brain MRI & Anatomy & RaTEScore & 21.77 [20.48--23.05] & -- & -- & 53.71 [52.25--55.19] & -- & 69.52 [68.36--70.67] & Hulu-Med-32B & +15.81 & 0.0001 & *** \\
Brain MRI & Brain MRI & Disease & Accuracy & 28.01 [25.52--30.65] & -- & -- & 46.59 [44.79--48.41] & -- & 63.85 [62.09--65.57] & Hulu-Med-32B & +17.25 & $5.29\times 10^{-46}$ & *** \\
Brain MRI & Brain MRI & Disease & RaTEScore & 16.48 [15.69--17.28] & -- & -- & 50.01 [49.11--50.95] & -- & 60.83 [60.03--61.61] & Hulu-Med-32B & +10.82 & 0.0001 & *** \\
US & CURG-US & Abnormality & Accuracy & 24.05 [21.25--27.10] & -- & 44.60 [42.62--46.61] & 42.06 [40.32--43.82] & 50.25 [48.24--52.27] & 62.21 [60.48--63.91] & Lingshu-7B & +11.96 & $3.52\times 10^{-26}$ & *** \\
US & CURG-US & Abnormality & RaTEScore & 25.58 [24.54--26.61] & -- & 54.29 [53.56--55.01] & 57.24 [56.44--58.04] & 53.19 [52.48--53.90] & 67.88 [67.01--68.77] & Hulu-Med-32B & +10.63 & 0.0001 & *** \\
US & CURG-US & Anatomy & Accuracy & 27.72 [24.94--30.69] & -- & 47.67 [45.77--49.57] & 44.39 [42.51--46.29] & 55.35 [53.45--57.24] & 62.73 [60.87--64.55] & Lingshu-7B & +7.38 & $1.64\times 10^{-11}$ & *** \\
US & CURG-US & Anatomy & RaTEScore & 24.03 [23.09--25.04] & -- & 54.64 [53.96--55.30] & 58.51 [57.63--59.38] & 53.84 [53.16--54.54] & 68.97 [68.00--69.91] & Hulu-Med-32B & +10.46 & 0.0001 & *** \\
US & CURG-US & Disease & Accuracy & 20.30 [15.91--25.54] & -- & 36.30 [33.10--39.62] & 38.54 [35.29--41.89] & 43.80 [40.44--47.21] & 49.94 [46.55--53.33] & Lingshu-7B & +6.14 & 0.00185 & ** \\
US & CURG-US & Disease & RaTEScore & 22.16 [20.50--23.85] & -- & 49.28 [47.89--50.63] & 50.77 [49.23--52.30] & 47.76 [46.40--49.08] & 57.62 [55.88--59.36] & Hulu-Med-32B & +6.86 & 0.0001 & *** \\
\bottomrule
\end{tabular}%
}
\suppTableBottomSpace
\begin{minipage}{\textwidth}\footnotesize\emph{Note.} Accuracy is used for closed-ended questions; RaTEScore is used for open-ended comparative interpretation.\end{minipage}
\end{table*}

\begin{table*}[!htbp]
\centering
\scriptsize
\setlength{\tabcolsep}{2.5pt}
\renewcommand{\arraystretch}{1.10}
\caption{External and independent public generative comparative interpretation performance (Figure~4b).}
\label{tab:supp_comparative_external_public_compact}
\suppTableTopSpace
\resizebox{\textwidth}{!}{%
\begin{tabular}{lllllllllllll}
\toprule
Benchmark & Dataset & Metric & RadFM & CheXagent & MedGemma & Hulu-Med & Lingshu & \ours{} & Best baseline & $\Delta$ & $p$ & Sig. \\
\midrule
External CXR & CheXpert Plus & Accuracy & 30.40 [27.70--33.10] & 37.30 [35.50--39.20] & 41.50 [39.80--43.30] & 44.50 [41.10--47.90] & 47.10 [44.00--50.20] & 64.70 [63.10--66.30] & Lingshu & +17.60 & $3.23\times 10^{-5}$ & *** \\
External CT & BIMCV-R & Accuracy & 30.10 [28.40--31.80] & -- & -- & 42.60 [41.40--43.80] & -- & 60.60 [59.40--61.80] & Hulu-Med & +18.00 & $3.06\times 10^{-12}$ & *** \\
Public VQA & MMXU & Accuracy & -- & 37.80 [36.10--39.60] & 49.10 [47.40--50.90] & 57.20 [55.50--59.00] & 64.60 [62.80--66.30] & 87.10 [85.80--88.20] & Lingshu & +22.50 & $1.48\times 10^{-108}$ & *** \\
External CXR & CheXpert Plus & RaTEScore & 20.10 [19.10--21.00] & 38.50 [37.30--39.60] & 54.10 [53.30--54.80] & 55.90 [54.10--57.70] & 54.70 [52.90--56.50] & 60.30 [59.30--61.30] & Hulu-Med & +4.40 & 0.00012 & *** \\
External CT & BIMCV-R & RaTEScore & 20.20 [19.70--20.70] & -- & -- & 51.10 [50.60--51.70] & -- & 57.90 [57.30--58.50] & Hulu-Med & +6.80 & 0.0001 & *** \\
Public VQA & Medical-Diff-VQA & RaTEScore & -- & 31.00 [30.80--31.30] & 29.10 [28.80--29.50] & 25.40 [25.20--25.60] & 23.10 [22.90--23.30] & 44.20 [43.90--44.50] & CheXagent & +13.20 & 0.0001 & *** \\
\bottomrule
\end{tabular}%
}
\suppTableBottomSpace
\begin{minipage}{\textwidth}\footnotesize\emph{Note.} Public VQA accuracy is evaluated on MMXU, whereas public VQA RaTEScore is evaluated on Medical-Diff-VQA.\end{minipage}
\end{table*}

\begin{table*}[!htbp]
\centering
\scriptsize
\setlength{\tabcolsep}{2.5pt}
\renewcommand{\arraystretch}{1.10}
\caption{Longitudinal follow-up generative comparative interpretation accuracy (Figure~5a). Groups with fewer than 15 evaluable cases are excluded (consistent with the figure).}
\label{tab:supp_longitudinal_compact}
\suppTableTopSpace
\resizebox{\textwidth}{!}{%
\begin{tabular}{lllrrrrlr}
\toprule
Level & Modality & Finding / attribute & CheXagent & Lingshu & Hulu-Med & \ours{} & Best baseline & $\Delta$ \\
\midrule
Finding & Chest X-ray & Pulmonary GGO & 56.54 & 28.50 & 22.90 & 71.03 & CheXagent-8B & +14.49 \\
Finding & Chest X-ray & Pulmonary Consolidation & 32.50 & 34.17 & 20.00 & 60.83 & CheXagent-8B & +28.33 \\
Finding & CT & Pulmonary GGO & -- & -- & 50.00 & 65.08 & Hulu-Med-32B & +15.08 \\
Finding & CT & Pulmonary Consolidation & -- & -- & 44.26 & 68.03 & Hulu-Med-32B & +23.77 \\
Finding & CT & Liver: Soft tissue density & -- & -- & 41.86 & 69.77 & Hulu-Med-32B & +27.91 \\
Finding & CT & Liver: Decreased density & -- & -- & 51.35 & 67.57 & Hulu-Med-32B & +16.22 \\
Finding & CT & Abd./Pelvic: Lymph node & -- & -- & 26.09 & 39.13 & Hulu-Med-32B & +13.04 \\
Attribute & Chest X-ray & Relative signal/density/intensity & 57.06 & 21.47 & 15.34 & 72.39 & CheXagent-8B & +15.34 \\
Attribute & Chest X-ray & Anatomical location & 51.16 & 69.77 & 55.81 & 97.67 & CheXagent-8B & +46.51 \\
Attribute & Chest X-ray & Spatial distribution & 55.56 & 46.03 & 30.16 & 80.95 & CheXagent-8B & +25.40 \\
Attribute & CT & Relative signal/density/intensity & -- & -- & 47.97 & 69.92 & Hulu-Med-32B & +21.95 \\
Attribute & CT & Anatomical location & -- & -- & 56.25 & 67.50 & Hulu-Med-32B & +11.25 \\
Attribute & CT & Spatial distribution & -- & -- & 40.00 & 62.50 & Hulu-Med-32B & +22.50 \\
Attribute & CT & Lesion size & -- & -- & 34.00 & 36.00 & Hulu-Med-32B & +2.00 \\
Attribute & CT & Internal composition & -- & -- & 47.37 & 92.11 & Hulu-Med-32B & +44.74 \\
\bottomrule
\end{tabular}%
}
\suppTableBottomSpace
\begin{minipage}{\textwidth}\footnotesize\emph{Note.} Values are accuracy (\%). GGO denotes ground-glass opacity. $\Delta$ is \ours{} minus the strongest modality-level baseline (CheXagent-8B for chest X-ray, Hulu-Med-32B for CT). Subgroups with $n<15$ evaluable cases—chest X-ray \emph{Lesion size}, \emph{Internal composition}, \emph{Margins}, \emph{Number of lesions}; CT finding \emph{Abd./Pelvic: Soft tissue}; CT attributes \emph{Margins}, \emph{Number of lesions}—are omitted to match Figure~5a.\end{minipage}
\end{table*}

\begin{table*}[!htbp]
\centering
\scriptsize
\setlength{\tabcolsep}{2.5pt}
\renewcommand{\arraystretch}{1.10}
\caption{Ablation results for model design choices (Figure~6), with 95\% confidence intervals and dataset-level significance.}
\label{tab:supp_ablation_compact}
\suppTableTopSpace
\resizebox{\textwidth}{!}{%
\begin{tabular}{lllccr c}
\toprule
Task & Modality & Entity & Variant (95\% CI) & Full model (95\% CI) & $\Delta$ & $p$ \\
\midrule
\multirow{12}{*}{Retrieval}
 & \multirow{3}{*}{CXR}   & Anatomy     & 63.84 {\tiny[62.67,\,65.00]} & 70.21 {\tiny[69.08,\,71.30]} & $+6.36$ & \multirow{3}{*}{$4.5{\times}10^{-84}$} \\
 &                        & Abnormality & 39.29 {\tiny[37.90,\,40.70]} & 48.44 {\tiny[47.01,\,49.87]} & $+9.15$ & \\
 &                        & Disease     & 63.77 {\tiny[62.05,\,65.46]} & 70.85 {\tiny[69.20,\,72.43]} & $+7.07$ & \\
\cmidrule(lr){2-7}
 & \multirow{3}{*}{CT}    & Anatomy     & 65.46 {\tiny[64.05,\,66.84]} & 69.39 {\tiny[68.02,\,70.73]} & $+3.93$ & \multirow{3}{*}{$2.5{\times}10^{-31}$} \\
 &                        & Abnormality & 46.89 {\tiny[45.59,\,48.19]} & 51.38 {\tiny[50.08,\,52.68]} & $+4.49$ & \\
 &                        & Disease     & 60.07 {\tiny[58.51,\,61.61]} & 65.67 {\tiny[64.15,\,67.15]} & $+5.60$ & \\
\cmidrule(lr){2-7}
 & \multirow{3}{*}{BRAIN} & Anatomy     & 58.15 {\tiny[54.89,\,61.34]} & 63.62 {\tiny[60.41,\,66.70]} & $+5.47$ & \multirow{3}{*}{$1.1{\times}10^{-16}$} \\
 &                        & Abnormality & 31.81 {\tiny[30.38,\,33.26]} & 36.09 {\tiny[34.62,\,37.59]} & $+4.28$ & \\
 &                        & Disease     & 72.69 {\tiny[71.19,\,74.14]} & 77.07 {\tiny[75.65,\,78.43]} & $+4.38$ & \\
\cmidrule(lr){2-7}
 & \multirow{3}{*}{US}    & Anatomy     & 67.76 {\tiny[64.94,\,70.46]} & 70.58 {\tiny[67.82,\,73.20]} & $+2.82$ & \multirow{3}{*}{$2.3{\times}10^{-6}$} \\
 &                        & Abnormality & 48.80 {\tiny[45.94,\,51.67]} & 52.65 {\tiny[49.79,\,55.50]} & $+3.85$ & \\
 &                        & Disease     & 76.62 {\tiny[69.34,\,82.61]} & 81.17 {\tiny[74.26,\,86.56]} & $+4.55$ & \\
\midrule
\multirow{12}{*}{\shortstack[l]{Comparative\\VQA}}
 & \multirow{3}{*}{CXR}   & Anatomy     & 70.64 {\tiny[67.72,\,73.39]} & 72.45 {\tiny[69.59,\,75.14]} & $+1.81$ & \multirow{3}{*}{$2.9{\times}10^{-8}$} \\
 &                        & Abnormality & 69.33 {\tiny[68.45,\,70.19]} & 70.95 {\tiny[70.09,\,71.80]} & $+1.62$ & \\
 &                        & Disease     & 60.49 {\tiny[58.92,\,62.04]} & 62.00 {\tiny[60.44,\,63.54]} & $+1.51$ & \\
\cmidrule(lr){2-7}
 & \multirow{3}{*}{CT}    & Anatomy     & 58.57 {\tiny[56.46,\,60.65]} & 62.34 {\tiny[60.25,\,64.37]} & $+3.77$ & \multirow{3}{*}{$6.1{\times}10^{-4}$} \\
 &                        & Abnormality & 68.65 {\tiny[67.49,\,69.79]} & 69.32 {\tiny[68.17,\,70.46]} & $+0.67$ & \\
 &                        & Disease     & 52.11 {\tiny[49.65,\,54.55]} & 53.05 {\tiny[50.59,\,55.49]} & $+0.94$ & \\
\cmidrule(lr){2-7}
 & \multirow{3}{*}{BRAIN} & Anatomy     & 62.29 {\tiny[59.57,\,64.94]} & 68.29 {\tiny[65.66,\,70.82]} & $+6.00$ & \multirow{3}{*}{$1.1{\times}10^{-18}$} \\
 &                        & Abnormality & 59.74 {\tiny[58.19,\,61.27]} & 63.58 {\tiny[62.06,\,65.08]} & $+3.84$ & \\
 &                        & Disease     & 59.88 {\tiny[58.09,\,61.64]} & 63.85 {\tiny[62.09,\,65.57]} & $+3.97$ & \\
\cmidrule(lr){2-7}
 & \multirow{3}{*}{US}    & Anatomy     & 60.73 {\tiny[58.86,\,62.57]} & 62.73 {\tiny[60.87,\,64.55]} & $+2.00$ & \multirow{3}{*}{$2.2{\times}10^{-5}$} \\
 &                        & Abnormality & 59.28 {\tiny[57.53,\,61.00]} & 62.21 {\tiny[60.48,\,63.91]} & $+2.93$ & \\
 &                        & Disease     & 51.14 {\tiny[47.75,\,54.52]} & 49.94 {\tiny[46.55,\,53.33]} & $-1.20$ & \\
\bottomrule
\end{tabular}%
}
\suppTableBottomSpace
\begin{minipage}{\textwidth}\footnotesize\emph{Note.} Retrieval values are Recall@1; comparative VQA values are closed-ended accuracy. The comparison is Shared-ViT (Variant) vs.\ MoE-ViT (Full) for retrieval, and InfoNCE-only (Variant) vs.\ Full warm-up+triplet (Full) for VQA. Bracketed ranges are 95\% Wilson score confidence intervals on the per-condition metric. The reported $p$ is a single dataset-level two-sided McNemar test pooled over all three conditions (one value per modality, not per condition); all are $<0.001$.\end{minipage}
\end{table*}

\end{document}